\documentclass[]{interact}

\usepackage{epstopdf}
\usepackage[caption=false]{subfig}

\usepackage[numbers,sort&compress]{natbib}
\bibpunct[, ]{[}{]}{,}{n}{,}{,}

\usepackage{amsmath,amssymb}
\usepackage{times}
\usepackage{graphicx}
\usepackage{color}
\usepackage{comment}
\usepackage{multirow, makecell}
\usepackage{newtxtext,newtxmath}
\usepackage{csquotes} 
\usepackage{rotating}
\usepackage{bbm}
\usepackage{latexsym}
\usepackage{algorithm}
\usepackage{algpseudocode}
\usepackage{url,hyperref}
\usepackage{anyfontsize}

\newcommand{\hcolor}{black}

\renewcommand\bibfont{\fontsize{10}{12}\selectfont}
\makeatletter
\def\NAT@def@citea{\def\@citea{\NAT@separator}}
\makeatother

\theoremstyle{plain}

\theoremstyle{definition}

\theoremstyle{remark}

\begin{document}

\articletype{Advanced Robotics Full Papers}

\title{Phase-Amplitude Reduction-Based Imitation Learning}

\author{
\name{Satoshi Yamamori\textsuperscript{a}\thanks{
CONTACT Satoshi Yamamori. Email: yamamori@atr.jp \\
This is an Accepted Manuscript of an article published by Taylor \& Francis in Advanced Robotics on 17 Dec 2024, available at: \url{https://doi.org/10.1080/01691864.2024.2441242}. \\
The implemented code is available: \href{https://github.com/LMGroup-KyotoU/phase-amplitude-reduction-based-imitation-learning}{GitHub repository}.
} 
and Jun Morimoto\textsuperscript{a, b}}
\affil{\textsuperscript{a}Department of Brain Robot Interface, ATR Computational Neuroscience Laboratories, Kyoto, Japan;
\textsuperscript{b}Graduate School of Informatics, Kyoto University, Kyoto, Japan}
}   

\maketitle

\begin{abstract}
In this study, we propose the use of the phase-amplitude reduction method to construct an imitation learning framework. Imitating human movement trajectories is recognized as a promising strategy for generating a range of human-like robot movements. Unlike previous dynamical system-based imitation learning approaches, our proposed method allows the robot not only to imitate a limit cycle trajectory but also to replicate the transient movement from the initial or disturbed state to the limit cycle. Consequently, our method offers a safer imitation learning approach that avoids generating unpredictable motions immediately after disturbances or from a specified initial state. We first validated our proposed method by reconstructing a simple limit-cycle attractor. We then compared the proposed approach with a conventional method on a lemniscate trajectory tracking task with a simulated robot arm. Our findings confirm that our proposed method can more accurately generate transient movements to converge on a target periodic attractor compared to the previous standard approach. Subsequently, we applied our method to a real robot arm to imitate periodic human movements. 
\end{abstract}


\begin{figure}
    \centering
    \includegraphics[width=\linewidth]{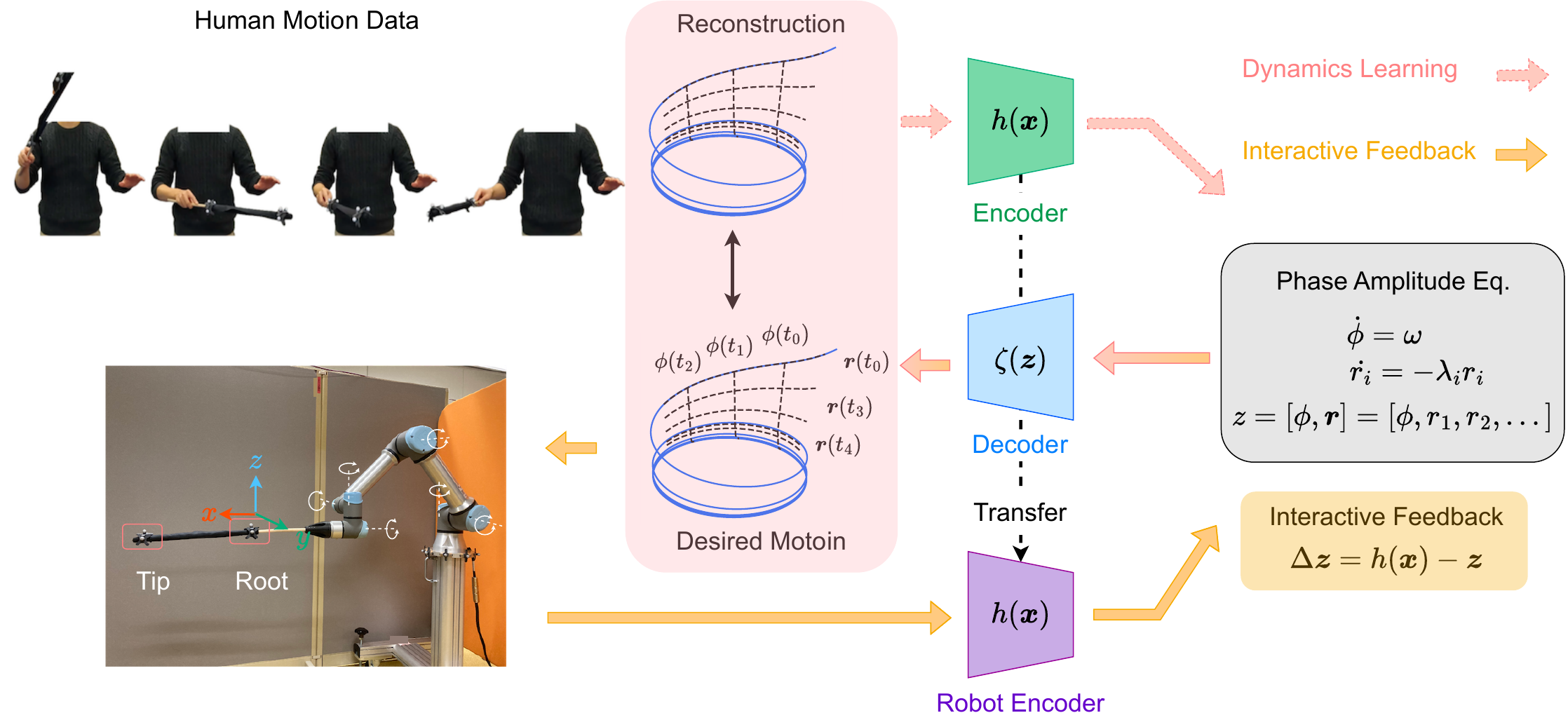}
    \caption{Phase-amplitude reduction-based imitation learning method. 
        Proposed method reconstructs a human trajectory, recorded by an optical motion capture system, using an encoder $h$ and a decoder $\zeta$. 
        Human trajectory is encoded in a latent space that follows the phase-amplitude equation representing a linear system with a frequency term $\omega$ and an exponent term $\lambda_i$, which determines the trajectory amplitude $r_i$ and phase $\phi$. 
        Encoder $h$ projects the robot state to latent space.
        Feedback connection is provided to regulate the latent variables according to the robot state.
    }
    \label{fig:concept}
\end{figure}

\section{Introduction}
In imitation learning approaches, a robot learns from human demonstration data to generate human-like motions \cite{schaal1996,ravichandar2020,hussein2017}. Studies in imitation learning have shown that robots can imitate human expert movements, such as kendama (cup-and-ball) \cite{miyamoto1996} and table tennis \cite{mulling2013}, skills that even humans need to acquire proficiency.
Recent advancements in deep learning have led to improvements in imitation learning performance \cite{pahic2020,chi2023diffusionpolicy,florence2021}, enabling flexible object manipulation \cite{yang2017}.
However, collecting large-scale human movement data to generate a wide variety of motions demands significant effort.
Thus, approximating the time series of human movements with a dynamical system to generate stable trajectories, even in cases where sufficient data is unavailable, can prove to be a valuable and extensively studied approach \cite{ijspeert2013,saveriano2021dynamic}.
Conventional dynamical movement primitive (DMP) approaches are limited since they can accurately approximate human movement only around the limit cycle and cannot adequately represent transient movements.
As a result, the reproduced robot behavior deviates from the demonstration data in a transient phase.

This study proposes a novel imitation learning framework that uses the phase-amplitude reduction method. The phase-amplitude reduction theory, developed based on the phase-reduction theory \cite{kuramoto2003chemical}, analyzes nonlinear oscillators and embeds the system state into a low-dimensional space \cite{Shirasaka2020}.
Each subspace of the embedded latent space corresponds to a phase of a limit cycle and an amplitude component converging to the limit cycle, making it ideal for analyzing dynamic motion.
Unlike the previous dynamical system-based imitation learning approach, the proposed method enables the robot to imitate not only a stable orbit but also the transient movement from the initial or disturbed state to the stable orbit.
Consequently, the proposed method provides a safer imitation learning method that does not generate unpredictable motions after disturbances or given an initial state. 
First, we verified our proposed method by reconstructing a simple limit-cycle attractor. 
Then, we applied the proposed method to a real robot arm to imitate periodic human movements. 
We confirmed that our proposed method could more accurately generate transient movements to converge on a target periodic attractor than the previous standard approach \cite{ijspeert2013, saveriano2021dynamic}.
As shown in Fig. \ref{fig:concept}, the proposed method uses a supervised learning framework based on variational inference to obtain an encoder-decoder that performs embedding and reconstruction from the demonstration data to the phase-amplitude space.

The proposed method comprises two components: 1) dynamics learning and 2) interactive feedback, aimed at assisting the robot in adhering to the desired trajectory obtained from human data. 
Besides the feedback from the desired dynamical system on the robot, the robot also provides its state feedback on the latent space through the encoder network.
The reduction in the phase-amplitude space reconstructs the transient dynamics of human motion while retaining the function to represent the limit cycle.

This study validated that the proposed method can learn a simple limit cycle model, perform a lemniscate curve-following task for a 6DoF arm robot, and replicate baton waving behavior exhibited by a human on a real robot. The results demonstrate that the proposed method can accurately reproduce the original dynamics, guide the robot toward the desired behavior, and successfully transition to a real robot.
The contributions of this study are as follows:
\begin{itemize}
    \item We proposed using the variational inference approach for stable learning of the encoder and decoder to imitate demonstrated movements.
    \item The proposed learning method was further equipped with the feedback mechanism from the robot to the latent dynamics extracted through the encoder-decoder learning to cope with external disturbances.
    \item The transient dynamics that represent the behaviors of states from initial or disturbed states to the limit cycle were successfully reconstructed in a simple example and also in a tracking task of a simulated arm robot.
    \item We confirmed that the real arm robot could imitate demonstrated human movement using the proposed learning method. 
\end{itemize}


The remainder of this paper is organized as follows. Section II introduces related studies. Section III explains the proposed method. Section IV introduces the experimental setups. Section V presents the results on both the simulation and the real system. Finally, Section VI concludes this paper.

\section{Related Works}

\subsection{Trajectory-based imitation learning}
Despite their simple mechanisms, the approaches that generate trajectories using dynamical systems, such as DMP, have demonstrated the ability to produce reliable motion. Model-based imitation learning facilitates reinforcement, inverse reinforcement, and imitation learning because smooth curves can connect the initial and target points \cite{schaal1999}. Additionally, the dynamics coupling approach between the real and target systems can modify systems that deviate from the desired dynamical system \cite{morimoto2008, Aljaz2018}. In this study, we construct a state feedback system on phase-amplitude space to enable trajectory tracking that considers the dynamic characteristics of the imitator.

\subsection{Latent representation in dynamics model}
This paper proposes a dynamic model-learning algorithm based on phase-amplitude reduction.
This dynamics model is suitable for imitation learning since the robot can explicitly select the dynamic mode of the target to follow. Nonlinear dynamics learning through latent space has evolved from subspace identification methods \cite{katayama2005} to include kernel methods \cite{katayama2005}, Koopman operators \cite{brunton2022}, Gaussian process regression \cite{lawrence2003, wang2005}, neural networks \cite{ghahramani1998}, methods restricted to stable dynamical systems \cite{khansari-zadeh2011, umlauft2017}, and mixed Gaussian distributions \cite{calinon2012}, among others. Notably, the phase-amplitude equation is classified as the Hammerstein-Wiener model, which sandwiches a linear dynamical system between static nonlinear functions. A similar approach for imitation learning involves neural networks and Koopman operators \cite{han2023on}.
Our proposed method endows the latent space structure with a distinctive dynamical property: phase and amplitude. This structure allows for the modulation of the desired dynamical system through the deceleration and acceleration of the phase, as well as the contraction and extension of the closed orbit via interactive feedback.

\section{Methods}

\subsection{Phase-amplitude Reduction Latent Dynamics}
As illustrated in Fig. \ref{fig:concept}, our proposed method enables a robot to learn the representation of a latent dynamical system capable of imitating human movement trajectories. Specifically, encoder and decoder networks were trained to represent the latent dynamical systems using phase-amplitude reduction. We employed the variational inference framework to train these networks. The encoder network takes physical observation space inputs, such as the end-effector position, orientation, and velocity.

\subsubsection{Phase-Amplitude Reduction}
In the proposed approach, we approximate a demonstrated human movement trajectory by a continuous-time dynamical system:
\begin{align}
    \dot{\boldsymbol{x}} = F(\boldsymbol{x}) \label{eq:auto-sys},
\end{align}
where $\boldsymbol{x}\in \mathbb{R}^N$ is the state variable and $F(\boldsymbol{x})$ denotes a vector field in the state space.
Assuming the autonomous dynamical system converges to a stable closed orbit, the nonlinear system can be reduced to the phase-amplitude equation \cite{Shirasaka2020} through the mapping $\boldsymbol{z}:=[\phi, \boldsymbol{r}] = h(\boldsymbol{x})$, where the dynamics of the latent variable $\boldsymbol{z}$ is defined as:
\begin{align}
    \dot{\boldsymbol{z}}&=f(\boldsymbol{z}) = \left [
        \omega, 
        -\boldsymbol{\lambda} \odot \boldsymbol{r}
    \right ],
    \label{eq:phase-amplitude-reduction}
\end{align}
where $f$ is the latent dynamics vector field, the latent variable $\boldsymbol{z}\in \mathbb{R}^{M}$ comprises the phase $\phi\in \mathbb{R}$ and amplitude $\boldsymbol{r}=[r_1,...,r_{M-1}]\in \mathbb{R}^{M-1}$ with the characteristic frequency as well as exponents $\omega>0$ and $\boldsymbol{\lambda}>0$, and the $\odot$ is the element-wise multiplication.
For the phased-amplitude dynamics, the following solution can be analytically derived
$\phi(t)=\phi(0) + \omega t, \boldsymbol{r}(t)= \exp(-\boldsymbol{\lambda} t) \odot \boldsymbol{r}(0)$;

The initial value set $\boldsymbol{x}(\phi)$, the isochron, is mapped onto the same phase $\phi$ of the limit cycle.
Similarly, the amplitude $\boldsymbol{r}$ represents the level set in the basin of the limit cycle attraction, the isostable.
In other words, the phase $\phi$ indicates the clock time on the limit cycle, and the amplitude $\boldsymbol{r}$ indicates the distance from the limit cycle. The function $h$ maps the state $\boldsymbol{x}$ to the phase-amplitude space $\boldsymbol{z}$.
Considering the latent space dimension is smaller than the system dimension, i.e., $M<N$, the latent dynamics representation can be referred to as a phase-amplitude "reduction".

If time $t$ is sufficiently large, with $\boldsymbol{\lambda}>0$, the each elements of amplitude $\boldsymbol{r}$ asymptotes to $0$.
Therefore, transient dynamics that are faster than the variable $\boldsymbol{z}$ can be ignored as they stay near the limit cycle for the timescale of the phase-amplitude system characterized by the inverse time constant $\boldsymbol{\lambda}$.
Therefore, observable $\boldsymbol{x}$ is reconstructed by learning a decoder $\hat{\boldsymbol{x}}=\zeta(\boldsymbol{z})$ from the latent variable $\boldsymbol{z}$ to observable $\boldsymbol{x}$.

This study aims to reconstruct the human demonstration data based on the phase-amplitude dynamics $f$ by learning the encoder $h$ and decoder $\zeta$, as shown in Fig. \ref{fig:concept}. The characteristic frequency $\omega$ and characteristic exponent $\boldsymbol{\lambda}$, the inverse time constant, can be estimated using time-series analysis, such as Fast Fourier Transform (FFT) or autocorrelation analysis.

Furthermore, the frequency $\omega$ and the exponent $\lambda$ exhibit invariance depending on $h$.
For the characteristic frequency, if $\phi'=h'(x):=ah(x)$ and $\zeta'(\phi'):=\zeta(\phi'/a)$, we obtain $\zeta'(h'(x))=\zeta(h(x))$, and $\dot{\phi'} =a\dot{\phi}= a\omega$, and the timescale of the latent space can be adjusted based on scaling of the observation function.
Additionally, for the characteristic exponent, by setting $r'=h'(x):=h(x)^a, \zeta'(r'):=\zeta(r'^{1/a})$, we also obtain $\zeta'(h'(x))=\zeta(h(x))$, and $\dot{r'}=ar^{a-1}\dot{r}=-a\lambda r'$.
Thus, the encoder has several representations for each different timescale in latent space to the extent that the above transformations are possible, and designing the exponents is robust. Therefore, the proposed algorithm can use pre-estimated exponents based on conventional spectrum analysis methods like FFT or autocorrelation analysis and reduce the learning cost of identifying dynamic components.

\subsubsection{Interactive Feedback}
\begin{figure}
    \centering
    \includegraphics[width=0.5\textwidth]{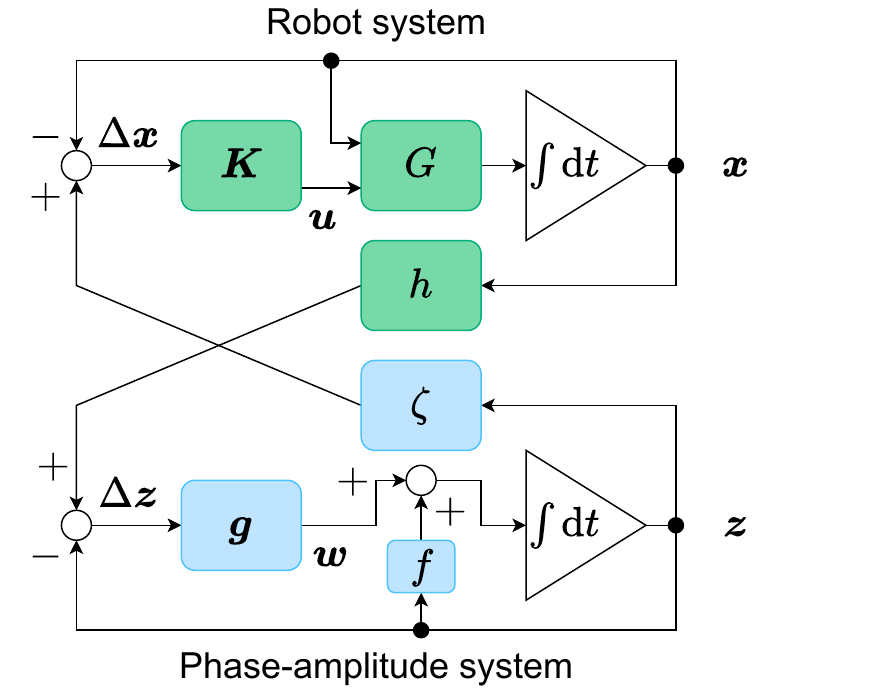}
    \caption{
        \hspace{-6pt}
        Interactive feedback system.
        Robot system $G$ and phase-amplitude system $f$ mutually feedback the errors 
        \textcolor{\hcolor}{$\Delta x$} and $\Delta z$.
    }
    \label{fig:interactive-feedback-block}
\end{figure}

Here, as suggested in \cite{morimoto2008}, we introduce a feedback controller by coupling the phase-amplitude dynamics with the observed sensor inputs (see also Fig. \ref{fig:concept}).
We introduced a weakly coupled system of feedback on the robot system $G$ with latent dynamical system $f$, described as:
\begin{align}
    \dot{\boldsymbol{x}} &= G(\boldsymbol{x}, \boldsymbol{u}), \boldsymbol{u}^T=\boldsymbol{K} \Delta \ \boldsymbol{x}^T , \\
    \dot{\boldsymbol{z}} &= f(\boldsymbol{z}) + \boldsymbol{w}, \boldsymbol{w}=\boldsymbol{g} \odot \Delta \boldsymbol{z},\\
    & \Delta \boldsymbol{x} = \zeta(\boldsymbol{z}) - \boldsymbol{x}, \\
    & \Delta \boldsymbol{z} = h(\boldsymbol{x}) - \boldsymbol{z} 
    \label{eq:feedback-system}
\end{align}
where $\boldsymbol{x}\in \mathbb{R}^N$ is the robot state variable, $\boldsymbol{u}\in \mathbb{R}^L$ and $\mathbf{K}\in \mathbb{R}^{L\times N}$ are the control input and gain, respectively. $\boldsymbol{g}\in \mathbb{R}^{M}$ is the feedback gain.
The encoder $h$ converts the robot state $\boldsymbol{x}$ into the latent variable $\boldsymbol{z}$, and the decoder $\zeta$ reconstructs the desired trajectory, as shown in Fig. \ref{fig:concept}.
Detailed implementation of the interactive feedback is depicted in Fig. \ref{fig:interactive-feedback-block}.


\subsection{Encoder-Decoder Learning}
\label{sec:Encoder-Decoder-Learning}
\begin{figure}
    \hspace{-0.05\columnwidth}
    \begin{minipage}[b]{0.65\linewidth}
        \subfloat[Multiple KL divergence.]{\includegraphics[width=1.0\linewidth]{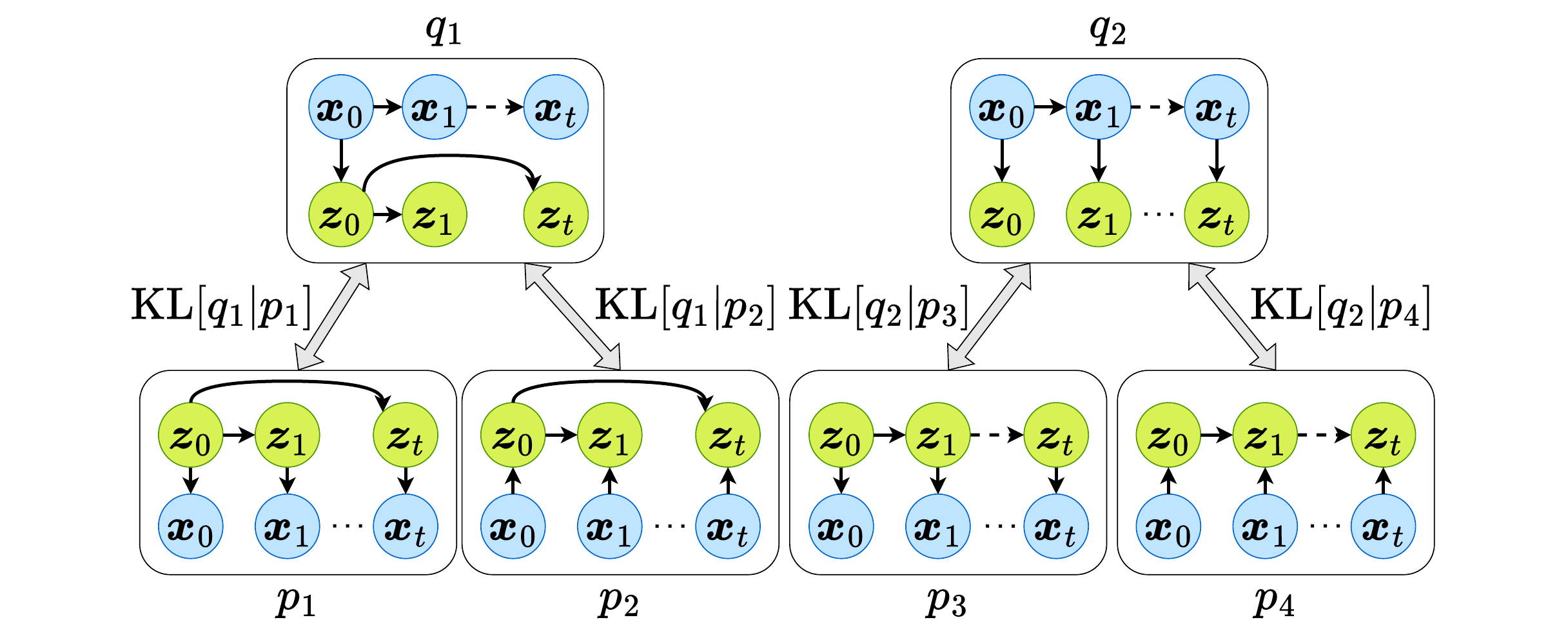}}
    \end{minipage}\hfill
    \begin{minipage}[b]{0.3\linewidth}
        \subfloat[Network architecture]{\includegraphics[width=0.99\linewidth]{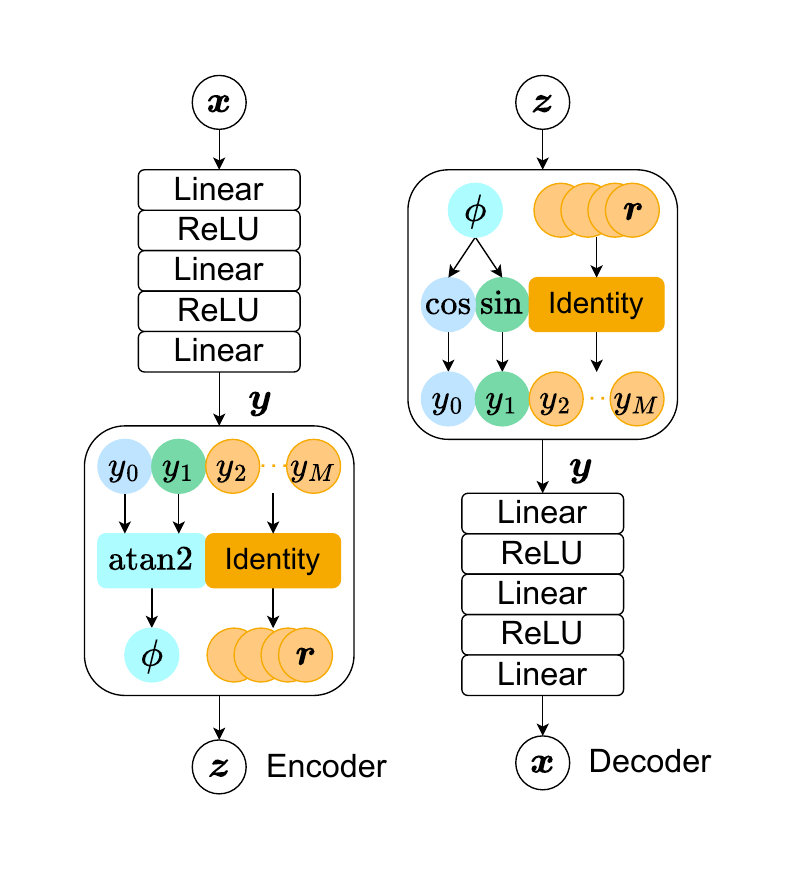}}
    \end{minipage}
    \caption{
        (a) Proposed probabilistic graphical models.
        Variational distributions $q_1$ and $q_2$ generate the latent variable $\boldsymbol{z}$ from observation $\boldsymbol{x}$. 
        Probabilistic distributions $p_1$, $p_2$, $p_3$, and $p_4$ represent the target probabilistic model. 
        We calculate four KL divergences to compare variational and probabilistic distributions. 
       $\mathrm{KL}[q_1|p_1]$ and $\mathrm{KL}[q_1|p_2]$ represent transient process losses through the time expansion in latent space. $\mathrm{KL}[q_2|p_3]$ and $\mathrm{KL}[q_2|p_3]$ represent the stationary process losses directly derived from the observation space.
        (b) Network architecture of the encoder $h$ and decoder $\zeta$.
        Encoder input and decoder output are provided through the two-layer ReLU networks and one linear layer.
        The state variable $\boldsymbol{x}$
        was encoded to the phase variable $\phi$ through using the inverse tangent $\mathrm{atan2}$.
        The other way around, the phase variable $\phi$ was decoded to the state variable $\boldsymbol{x}$ through using the sinusoidal functions $\sin$ and $\cos$.
        $\mathrm{KL}[q_1|p_1]$, $\mathrm{KL}[q_1|p_2]$, and $\mathrm{KL}[q_2|p_4]$ were used to stabilize the learning process.
        }
    \label{fig:model-and-network}
\end{figure}

The proposed method adopts the Kullback--Leibler (KL) divergence minimization based on variational inference. Similar to the $\beta$ variational autoencoder ($\beta$-VAE) \cite{higgins2017}, encoder $h$ and decoder $\zeta$ were learned for reconstructing the human trajectory $\boldsymbol{x}$ through latent $\boldsymbol{z}$ variables (Fig. \ref{fig:concept}). The variational inference approach estimates the posterior distribution $q(Z|X)$ and the probabilistic model distribution $p(Z)p(X|Z)$ given data distribution $p(X)$ by alternately updating these distributions to minimize $\mathrm{KL}[p(X)q(Z|X)\|p(Z)p(X|Z)]$.
The graphical model of data $\boldsymbol{x}$ and latent variables $\boldsymbol{z}$ can be measured using $\mathrm{KL}[q_1|p_1]$, as shown on the left side of Fig. \ref{fig:model-and-network} (a).
Here, we consider the $T$ length time series of observations $\mathbf{X}=\{\boldsymbol{x}_0, \boldsymbol{x}_1, ...,\boldsymbol{x}_{T}\}$ and latents $\mathbf{Z}=\{\boldsymbol{z}_0, \boldsymbol{z}_1, ..., \boldsymbol{z}_{T}\}$ and the following:
\begin{align}
    \min_{h,\zeta}&\ \mathrm{KL}[q|p], \label{eq:kl-objective}\\
    q(\tau)&=p(\mathbf{X})\cdot e(\boldsymbol{z}_0|\boldsymbol{x}_0)\cdot \prod_{k=1}^{T} m_1(\boldsymbol{z}_{k}|\boldsymbol{z}_0), \label{eq:variational-distribution}\\
    p(\tau)&=p(\boldsymbol{z}_0) \cdot d(\boldsymbol{x}_0|\boldsymbol{z}_0) \cdot \prod_{k=1}^{T} m_1(\boldsymbol{z}_{k}|\boldsymbol{z}_0)d(\boldsymbol{x}_k|\boldsymbol{z}_k) \label{eq:target-distribution},
\end{align}
where $\tau=\{\boldsymbol{x}_0,\boldsymbol{z}_0,...,\boldsymbol{x}_{T},\boldsymbol{z}_{T}\}$ is a trajectory of observation $\boldsymbol{x}$. $e,d$ and $m_1$ denote the probabilistic models of the encoder, decoder, and latent dynamical system. $p(\mathbf{X})$ denotes data distribution and $p(\boldsymbol{z}_0)$ is the prior.
To make the learning procedure robust against the outlier, we adopted the Laplace distribution
as $m_1(\boldsymbol{z}_{k}|\boldsymbol{z}_0)=\mathcal{L}(f(\boldsymbol{z}_0, k),b_f)$, $e(\boldsymbol{z}_k|\boldsymbol{x}_k) = \mathcal{L}(h(\boldsymbol{x}_k), b_h)$, $p(\boldsymbol{z}_0)=\mathcal{L}(0, b_0)$, and $d(\boldsymbol{x}_k|\boldsymbol{z}_k) = \mathcal{L}(\zeta(\boldsymbol{z}_k),1)$.
Particularly, scale parameters $b_f, b_h, b_0$ of Laplace distribution are fixed. Latent dynamics model $m_1(\boldsymbol{z}_{k}|\boldsymbol{z}_0)$ has a location parameter $f(\boldsymbol{z}_0, k)$ as a discretized dynamical system of Eq. (\ref{eq:phase-amplitude-reduction}):
\begin{eqnarray}
    f(\boldsymbol{z}_0, k)&=&[\phi_k, \boldsymbol{r}_k] \nonumber \\
       &=& \left [\phi_0 + \omega k\Delta T,
        \exp(-\boldsymbol{\lambda} k\Delta T) \odot \boldsymbol{r}_0 \right ], 
\end{eqnarray}    
where $\Delta T$ is the step size.
As defined in Eqs. (\ref{eq:variational-distribution}) and (\ref{eq:target-distribution}), the latent variable $z_k$ depends only on $z_0$ rather than $z_{k-1}$. In this modeling, we assumed that the robot system is close to a deterministic system, e.g., ridged body dynamics, and then we modeled the noise depending only on the initial state for simplicity. Furthermore, technically, using this model has the advantage of stabilizing the learning process of the neural network models because this approach can avoid the vanishing gradient problem.

The KL minimization problem in Eq. (\ref{eq:kl-objective}) can be transformed into an absolute error minimization problem (Eq. (\ref{eq:kl-ae-expantion})) since the Laplace distribution entropy is independent of the location parameters, and the distributions $q(\tau),p(\tau)$ share a common distribution $m_1$ as:
\begin{align}
    \mathrm{KL}[q|p] &= \mathbb{E}_q \left [ \log \frac{e(\boldsymbol{z}_0|\boldsymbol{x}_0)}{p(\boldsymbol{z}_0)} +
    \sum_{k=0}^{T} 
    -\log d(\boldsymbol{x}_k|\boldsymbol{z}_k) \right ] -H(\mathbf{X})\nonumber\\
    &= \mathbb{E}_q \left [
    \frac{\left |\boldsymbol{z}_0\right |}{b_2}  + 
    \sum_{k=0}^{T} \left |\boldsymbol{x}_{k} - \zeta(\boldsymbol{z}_k) \right | 
    \right ] \
    \underset{\mathrm{constant\ for}\ h\ \mathrm{and}\ \zeta}{\underline{-H(\mathbf{X}) - H(\boldsymbol{z}_0|\boldsymbol{x}_0) + \mathrm{const}}}\ , \label{eq:kl-ae-expantion}\\
    & \boldsymbol{z}_{k} = \begin{cases}
            h(\boldsymbol{x}_0) + \boldsymbol{\epsilon}_0 & \text{if } k = 0,\ \boldsymbol{\epsilon}_0 \sim \mathcal{L}(\boldsymbol{0}, b_h) \\
            f(\boldsymbol{z}_0, k) + \boldsymbol{\epsilon}_k & \text{otherwise}, \ \boldsymbol{\epsilon}_k \sim \mathcal{L}(\boldsymbol{0}, b_f)
            \end{cases},  \label{eq:m1-z-generate} 1\leq k\leq T
\end{align}
where $\mathbb{E}_{P(x)}[x]=\int_{\mathcal{X}} P(x)x\mathrm{d}x$ is an expectation of the distribution $P(x)$ and $H(X)=\mathbb{E}_{p(X)}[-\log p(X)]$ is entropy.
A detailed derivation is given in the Appendix.
This objective function can be minimized using the gradient descent method by employing the reparameterization trick \cite{higgins2017}. 


\subsection{Learning Algorithm}
We introduce the variational inference approach for efficient and stable learning of the encoder $h$ and decoder $\zeta$. 
The concrete algorithm is shown Alg. \ref{alg:latent-dyanmics} that incorporates the following ideas:
\begin{enumerate}
    \item Define two types of distributions allowing the algorithm to satisfy both stationary and more transient objectives. 
    
    \item Introduce a $\lambda$-scaled discount factor to decay the error on the limit cycle.
    Because the observed data $\mathbf{X}$ becomes close to a stationary distribution over time, the error decreases by the discount rate to improve the importance of the transient dynamics.
    
    \item 
    Because the phase range is $\phi \in (-\pi, \pi]$, the phase variable $\phi$ has a state jump from $\pm \pi$ to $\mp \pi$ and does not follow Laplace distribution. Therefore, 
    if the jump between consecutive angles is greater than $\pi$, we “unwrap” the phase variable, where the unwrap function shifts the variable by adding multiples of $\pm 2\pi$ until the jump is less than $\pi$.   
\end{enumerate}

\begin{algorithm}
    \caption{Learning Encoder / Decoder}\label{alg:latent-dyanmics}
    \begin{algorithmic}[1]
        \State {Demonstration Data: $\mathcal{D}$}
        \State Estimation of the exponents $\omega$ and $\boldsymbol{\lambda}$ using FFT.
        \State Hyper-Parameter: $\gamma$, $|B|$
        \State Learning network: $h,\zeta$
        \While{update times}
            \State Batch sample $\mathbf{B}\in \mathbb{R}^{|\mathbf{B}|\times T \times N}$ from $\mathcal{D}$
            \State Encoding $\boldsymbol{z}_k'\sim e(\boldsymbol{z}_k|\boldsymbol{x}_k),\boldsymbol{x}_k\sim \mathbf{B}$
            \State Rollout $\boldsymbol{z}_k \sim m_1(\boldsymbol{z}_k|\boldsymbol{z}_0)$
            \State Unwrap $\phi_k$ in $\boldsymbol{z}_k$ and $ \boldsymbol{z}_k'$
            \State Calculation Loss $L(h,\zeta;\mathbf{B})$
            \State Update Network for $h,\zeta$
        \EndWhile
    \end{algorithmic}
\end{algorithm}

\subsubsection{Probabilistic Models}
As shown in Fig. \ref{fig:model-and-network} (a), the following six probability distributions were defined:
\begin{align}
    q_1(\tau)&=p(\mathbf{X})\cdot e(\boldsymbol{z}_0|\boldsymbol{x}_0)\cdot \prod_{k=1}^{T} m_1(\boldsymbol{z}_k|\boldsymbol{z}_0), 
    \label{eq:val-prob-1}\\
    p_1(\tau)&=p(\boldsymbol{z}_0)\cdot d(\boldsymbol{x}_0|\boldsymbol{z}_0)\cdot\prod_{k=1}^{T} m_1(\boldsymbol{z}_k|\boldsymbol{z}_0)d(\boldsymbol{x}_k|\boldsymbol{z}_k),
    \label{eq:targ-prob-1}\\
    p_2(\tau)&=p(\mathbf{X})\cdot e(\boldsymbol{z}_0|\boldsymbol{x}_0)\cdot \prod_{k=1}^{T} m_2(\boldsymbol{z}_k|\boldsymbol{z}_0,\boldsymbol{x}_k),
    \label{eq:targ-prob-2}\\
    q_2(\tau)&=p(\mathbf{X})\cdot \prod_{k=0}^{T} e(\boldsymbol{z}_{k}|\boldsymbol{x}_k),
    \label{eq:val-prob-2}\\
    p_3(\tau)&=p(\boldsymbol{z}_0)\cdot d(\boldsymbol{x}_0|\boldsymbol{z}_0)\cdot \prod_{k=1}^{T} m_1'(\boldsymbol{z}_{k}|\boldsymbol{z}_{k-1})d(\boldsymbol{x}_k|\boldsymbol{z}_k),
    \label{eq:targ-prob-3}\\
    p_4(\tau)&=p(\mathbf{X})\cdot e(\boldsymbol{z}_0|\boldsymbol{x}_0)\cdot \prod_{k=1}^{T} m_2(\boldsymbol{z}_k|\boldsymbol{z}_{k-1},\boldsymbol{x}_k),
    \label{eq:targ-prob-4}
\end{align}
where $q_{\bullet}$ denotes the variational distribution and $p_{\bullet}$ denotes the model distribution, as defined in Section \ref{sec:Encoder-Decoder-Learning}.
Note that in data distribution $p(\mathbf{X})$, the stationary data $\boldsymbol{x}$ on the closed-orbit occupancy than transient data far from the closed-orbit since the transition part almost converges to the origin during the data collection.
Therefore, the batch sample for training is mainly sampled from stationary data.
Here, the probability distribution $q_1$ plays a role in training the long-term prediction. It sequentially generates $k$ steps future latent variable  $\boldsymbol{z}_k$ from the initial state $\boldsymbol{x}_0$. This process enhances the encoder $e(\boldsymbol{z}_0|\boldsymbol{x}_0)$ learning through the gradient backpropagation from future reconstruction loss, inducing sensitive learning against the transient dynamics.
Conversely, since distribution $q_2$ generates $\boldsymbol{z}_k$ from each time observation $\boldsymbol{x}_k$, the backpropagation is limited around the local timestep. Consequently, learning how to reconstruct batch samples, i.e., the neighborhood of the closed orbit, is enhanced. The latent dynamics model $m_1'(\boldsymbol{z}_{k}|\boldsymbol{z}_{k-1})=\mathcal{L}(f(\boldsymbol{z}_{k-1}, 1), 1)$ in Eq. (\ref{eq:targ-prob-3}) is redefined as a one-step prediction model.

The model $m_1$ is conditioned only by the initial latent variable $\boldsymbol{z}_0$; 
however, model $m_2$ is also conditioned by the observation $\boldsymbol{x}_k$ at that time as:
\begin{align}
    m_2(\boldsymbol{z}_k|\boldsymbol{z}_0,\boldsymbol{x}_k) &= \mathcal{L}\left (
    f(\boldsymbol{z}_0, k, \boldsymbol{x}_k),1 \right ),\\
    f(\boldsymbol{z}_0, k, \boldsymbol{x}_k) &= f(\boldsymbol{z}_0, k) + \kappa (h(\boldsymbol{x}_k) - f(\boldsymbol{z}_0, k)),
    \label{eq:m2-integral-equation}
\end{align}
where $\kappa  \in [0, 1]$ is the intensity parameter.
The model $m_2$ matches the transient dynamics of latent variable $\boldsymbol{z}_k$ with the encoder function $h$ of observation $\boldsymbol{x}_k$.
The difference between the models $p_1$ and $p_2$ lies in the direction of the conditions (see directions of arrows in Fig. \ref{fig:model-and-network} (a)).
Model $p_1$ represents the decoding probability from $\boldsymbol{z}$ to $\boldsymbol{x}$, whereas model $p_2$ represents the filtering probability from $\boldsymbol{x}$ to $\boldsymbol{z}$.

\subsubsection{Scaled Absolute Error Minimization}
We derived the six absolute error loss functions from the four KL divergences $\mathrm{KL}[q_1|p_1]$, $\mathrm{KL}[q_1|p_2]$, $\mathrm{KL}[q_2|p_3]$, and $\mathrm{KL}[q_1|p_4]$ with two additional losses regarding the difference in the observable $\boldsymbol{x}$, as shown below:
\begin{align}
    L(h,\zeta;\mathbf{X})&= L_\mathrm{Rec} + L_\mathrm{Enc} +
    L_\mathrm{Dec} + L_\mathrm{Lat} \nonumber \\
    &+ \sqrt{\Delta T} \cdot L_{\mathrm{Rec},\mathrm{Diff}} + \sqrt{\Delta T} \cdot L_{\mathrm{Dec},\mathrm{Diff}},\\
     L_\mathrm{Rec} &= \mathbb{E}_{q_1} \left [(1-\gamma)\sum_{k=0}^{T} \gamma^k\left|\boldsymbol{x}_k - \zeta(\boldsymbol{z}_k)\right|  \right],\\
     L_\mathrm{Enc} &=\mathbb{E}_{q_1} \left [
    (1-\gamma) \sum_{k=1}^{T} \gamma^k \kappa \left | \boldsymbol{c}_k \odot (h(\boldsymbol{x}_k) - f(\boldsymbol{z}_0, k)) \right |
\right], \\
    L_\mathrm{Dec} &= \mathbb{E}_{q_2} \left [
    (1-\gamma)\sum_{k=0}^{T} \gamma^k|\boldsymbol{x}_k - \zeta(\boldsymbol{z}_k')|
    \right], \\
    L_\mathrm{Lat} &= \mathbb{E}_{q_2} \left [
        (1-\gamma) \sum_{k=1}^{T} \gamma^k(2-\kappa)|\boldsymbol{c}_k\odot(h(\boldsymbol{x}_k) - f(\boldsymbol{z}_{k-1}', 1))|
    \right], \\
    L_\mathrm{Rec,Diff} &= \mathbb{E}_{q_1} \left [
        (1-\gamma)  \sum_{k=0}^{T-1}  \gamma^k  
 |\Delta(\boldsymbol{x}_{k+1}, \boldsymbol{x}_k) - \Delta(\zeta(\boldsymbol{z}_{k+1}), \zeta(\boldsymbol{z}_k))| 
    \right],\\
    L_\mathrm{Dec,Diff} &= \mathbb{E}_{q_2} \left [
        (1-\gamma)   \sum_{k=0}^{T-1} \gamma^k  |\Delta(\boldsymbol{x}_{k+1}, \boldsymbol{x}_k) - \Delta(\zeta(\boldsymbol{z}_{k+1}'), \zeta(\boldsymbol{z}_k'))| 
    \right],\\
    &\boldsymbol{z}_k \mathrm{\ is\ sampled\ from\ Eq.\ (\ref{eq:m1-z-generate})} \nonumber, \\
    &\boldsymbol{z}_k' = h(\boldsymbol{x}_k) + \boldsymbol{\epsilon}_k, \boldsymbol{\epsilon}_k \sim \mathcal{L}(0, b) \nonumber, \\
    &\Delta(\boldsymbol{x}_{k+1},\boldsymbol{x}_k) = \frac{\boldsymbol{x}_{k+1} - \boldsymbol{x}_k}{\Delta T} \nonumber,\\
    & \boldsymbol{c}_k = \frac{1 - \gamma \boldsymbol{d}}{1 - \gamma}\odot \boldsymbol{d}^k,\  
    \boldsymbol{d} = [1, \exp(-\boldsymbol{\lambda}\Delta T)]\in \mathbb{R}^M
    \nonumber,
\end{align}
where $\odot$ represents element-wise multiplication, i.e., Hadamard product.
Latent variables $\boldsymbol{z}_k$ and $\boldsymbol{z}'_k$ are generated by $q_1$ and $q_2$, respectively, for Monte Carlo integration of the KL divergences.
The latent variable $\boldsymbol{z}$ is a random variable that transitions through the phase-amplitude Eq. (\ref{eq:m1-z-generate}). In contrast, $\boldsymbol{z}'_k$ is a time-invariant latent variable generated only by the encoder $h(\boldsymbol{x}_k)$.
These loss terms are derived from KL divergence terms as described in the Appendix, and the terms have the following correspondence, respectively: $\mathrm{KL}[q_1|p_1] \rightarrow L_\mathrm{Rec}$; $\mathrm{KL}[q_1|p_2] \rightarrow L_\mathrm{Enc}$; $\mathrm{KL}[q_2|p_3] \rightarrow L_\mathrm{Dec}$ and $L_\mathrm{Lat}$; and $\mathrm{KL}[q_2|p_4] \rightarrow L_\mathrm{Lat}$
The loss functions based on $q_1$: $L_\mathrm{Rec}$ and $L_\mathrm{Dec}$ evaluate the long-term predictive performance of the noiseless latent variable. 
Conversely, the $q_2$-based loss functions $L_\mathrm{Dec}$ and $L_\mathrm{Lat}$ evaluate the one-step prediction performance based on the current state $\boldsymbol{x}$. 
The $L_\mathrm{Enc}$ and $L_\mathrm{Lat}$ evaluate the latent variable consistency between predicted and encoded.
Additionally, because the velocity component is more critical in the dynamics model, the two loss functions $L_\mathrm{Rec, Diff}$ and $L_\mathrm{Dec, Diff}$ are added as supplementary loss functions to $L_\mathrm{Rec}$ and $L_\mathrm{Diff}$ to enhance the long-term prediction performance further.
The timestep-scaled coefficient $\sqrt{\Delta T}$ works as weights for the loss functions $L_\mathrm{Rec,Diff}$ and $L_\mathrm{Dec,Diff}$. 
Furthermore, the error is computed by exponential weighted averaging of $\gamma$ and $\boldsymbol{c}_k$.
The discount factor $\gamma$ commonly occurs in loss functions.
The $\lambda$-scaled discount factor $\boldsymbol{c}_k$ decays the encoder error for each timestep $k$ toward the future.

\subsubsection{Structured Neural Network}
We modified the activation function of the two-layered perceptron, as shown in Fig. \ref{fig:model-and-network} (b).
The last layer activation function $\mathrm{atan2}$ converts encoder output $y_0$ and $ y_1$ into $\phi \in (-\pi, \pi)$.
The unwrapping function shown in line 9 in Alg. \ref{alg:latent-dyanmics}, activated value $\phi$ recovered the set of real numbers $\mathbb{R}$.
The decoder first layer converts contrary the $\phi\in\mathbb{R}$ into the circumference $S^1$ by activation functions $\sin$ and $\cos$.
The unwrapping process shifts the variable by adding multiples of $\pm 2\pi$ until the jump is less than $\pi$ to avoid discontinuity of the phase variables.

\section{Experimental Setups}
We conducted three experiments to confirm our contributions: 1) learning a dynamical system representation that can reproduce the transient dynamics of the demonstration data, 2) end-effector tracking control using a simulated arm robot, and 3) imitation learning from human movement data on a real arm robot.

\subsection{Simple Limit Cycle}
First, we validated whether the proposed learning frameworks can reconstruct the transient dynamics with a simple limit cycle \cite{martin1993,Strogatz1992}:
\begin{align}
    \dot{\boldsymbol{x}} &= \begin{bmatrix}0 & -\omega \\ \omega & 0 \end{bmatrix}  \boldsymbol{x} - (|\boldsymbol{x}|^2 - \alpha )\boldsymbol{x}, \label{eq:simple-limit-cycle}
\end{align}
where the state $\boldsymbol{x}=[x_1, x_2]\in \mathbb{R}^2$, $\alpha=1$ is the scale parameter and $\omega=2$ is the characteristic frequency of the limit cycle.
The state space and vector field is shown in Fig. \ref{fig:simple-limit-cycle-figs} (a).
Therefore, we can analytically derive the encoder function using the phase-amplitude equation with the observation function
$\phi = \mathrm{atan2}(x_2, x_1)$,
$r = \left (\frac{|\boldsymbol{x}|^2 - \alpha}{|\boldsymbol{x}|^2} \right )^{\frac{\lambda}{2 \alpha}}$:
\begin{align}    \label{eq:true-encoder}
    \dot{\phi} &= \omega, \\
    \dot{r} &= -\lambda r, \nonumber
\end{align}
where $\omega > 0$ and $\lambda > 0$ are the characteristic frequency and exponent, respectively. 
The exponent $\lambda$ can be defined with $\alpha$ as $\lambda=2 \alpha$.

The initial state of the training data is sampled a certain length away in the normal direction from a point on the limit cycle where the point is sampled from a uniform distribution of $[-\pi, \pi)$ and the length is sampled from a Gaussian distribution with mean $0$ and standard deviation $\sqrt{\alpha}$.
Here, $\sqrt{\alpha}$ means the limit cycle radius from the origin point.
The time step length in one trajectory is 100 steps, just the half cycles. 
We repeat the initial state sampling and numerical integration until collecting the size of data steps in Table \ref{tab:parameter-list}.


\subsection{Lemniscate tracking control}
Second, we evaluate the proposed method with the motion generation and feedback mechanisms by comparing our method with the dynamic movement primitives (DMPs) method \cite{ijspeert2013, Peternel2016} in transient situations.
Furthermore, we compared the reproducibility of the 6DOF single-arm robot end-effector movement in a lemniscate tracking task.

The demonstration data with the 6DoF arm robot tracking task were collected from a physics simulator, as shown in Fig. \ref{fig:ur5e-lemniscate} (a).
First, we solved the inverse kinematics problem and derived the target joint angles that correspond to the lemniscate end-effector trajectory. Then, we used a proportional and derivative (PD) servo control to track the target joint angles. The dynamics representation of the movement trajectories was learned from the demonstration data, and the observable $\boldsymbol{x}$ was reconstructed as shown in Table \ref{tab:observables}.
The characteristic frequencies were estimated using autocorrelation analysis. 
The characteristic exponents were obtained for 32 points equally divided on a logarithmic scale from the range listed in Table \ref{tab:parameter-list}.
The minimal and maximum characteristic exponents were selected from the spectrum norm range on the FFT frequency analysis.
Subsequently, the PD servo follows the trajectory generated by the learned latent system with interactive feedback.

\subsection{Human movement imitation}
Finally, we conducted a human movement imitation task using a real 6DoF arm robot as shown in Fig. \ref{fig:human-to-robot-result} (a) and (b).
A motion capture system measures the baton movement human swings, and our proposed method reconstructs the baton movement trajectory. 
Subsequently, the decoded state of the acquired latent dynamics was used to generate the robot arm movement trajectory.
The real arm robot tracked the generated position and orientation from the learned latent system in the hand coordinate system with the feedback connection with the marker position on the baton attached to the robot.
For the real robot experiment, we introduced a feedforward constant term $\boldsymbol{\phi}_\mathrm{ff}=[0.3,1.2]$ into the interactive feedback system to compensate for phase delay due to the tracking performance of the PD servo controller of the robot system as 
$\Delta \boldsymbol{\phi} = h_\phi(\boldsymbol{x}) - \boldsymbol{\phi} + \boldsymbol{\phi}_\mathrm{ff}$, here $h_\phi(\boldsymbol{x})$ is the phase elements in encoder output.

Table \ref{tab:observables} lists the observables to learn latent dynamics.
The motion-captured trajectories were the positions of the baton, which a human conducts with a 3-beat rhythm for 24 s. Its tempos are 30 times per 60 s.
The sampling rate of the motion capture system was 360 Hz, and the control period for the real robot arm UR5e was 50 Hz.
As a preprocessing step, the trajectories were downsampled from 360 Hz to 50 Hz and filtered by a low-pass filter with a cutoff frequency of $5$ Hz. 
Characteristic frequency and characteristic exponents were pre-estimated using autocorrelation analysis and FFT frequency analysis on human data.
The characteristic frequencies were determined by finding the two frequencies $\boldsymbol{\omega}=[\omega_0, \omega_1]$ from the autocorrelation analysis shown in Table \ref{tab:parameter-list}.
Unlike previous experiments, the data to be reconstructed has a large periodic orbit and a smaller periodic orbit divided into three parts, as shown in Fig. \ref{fig:human-to-robot-result} (c). 
Therefore, since the training data had a two-dimensional torus rather than a single limit cycle, the experiment was carried out with two-phase variables in the latent variable.
We set the characteristic exponents as 30 by equally dividing the logarithmic scale between a range of minimum and maximum values provided in Table \ref{tab:parameter-list}.
The maximum characteristic exponent was set at 1.5 Hz, where the spectral norm is sufficiently small due to the frequency analysis. The minimum characteristic exponent was set based on the higher characteristic frequency $\omega_1=0.51$ Hz. Because the time scale in latent variable z is different for each characteristic frequencies $\boldsymbol{\omega}$ and exponents $\boldsymbol{\lambda}$, the feedback gain $\boldsymbol{g}$ was modified to match the scale of the characteristics. That is, $\boldsymbol{g} = g / \omega_0 [\boldsymbol{\omega}, \boldsymbol{\lambda}]$, here, $\omega_0$ is the slowest characteristic frequency. 

Furthermore, our control system is a CPU-only PC (Intel Core i7-3770K CPU @ 3.5 GHz, Memory 20 GB) without GPU because the integral process to predict the future state becomes redundant owing to the use of the analytical solution of the latent dynamical system.

\section{Results}
\subsection{Simple Limit Cycle}
\begin{figure}
    \centering
    \begin{minipage}[b]{0.3\textwidth}
        \centering
        \subfloat[Target limit cycler attractor.]{
        \includegraphics[width=0.9\linewidth]{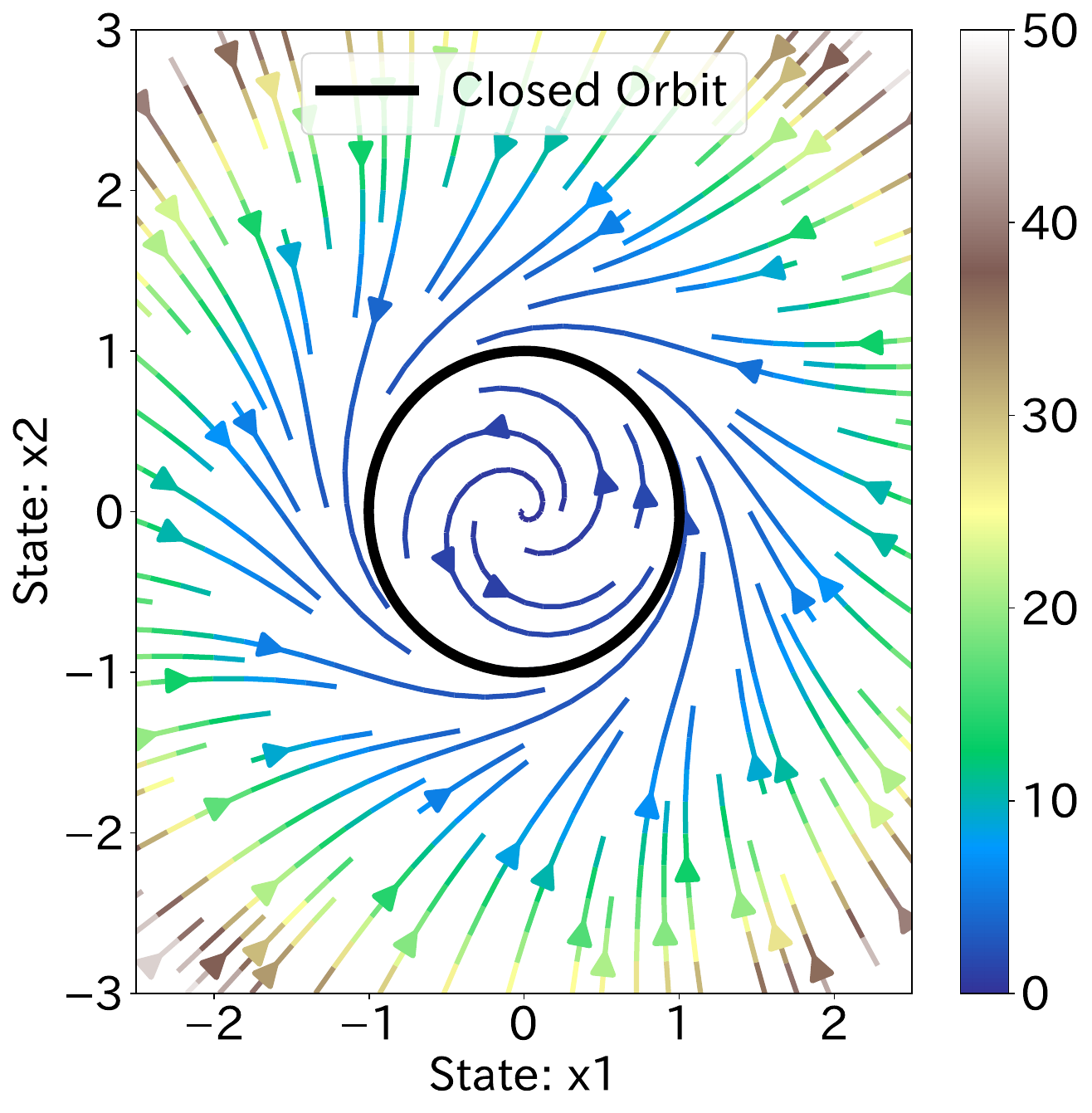}}
    \end{minipage}
    \begin{minipage}[b]{0.3\textwidth}
        \centering
        \subfloat[Different data size.]{\includegraphics[width=0.9\linewidth]{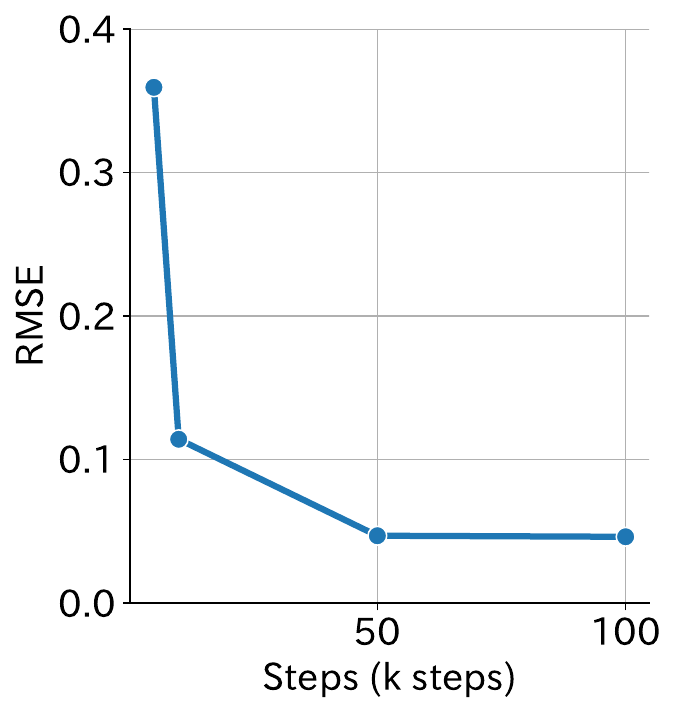}}
    \end{minipage}
    \begin{minipage}[b]{0.3\textwidth}
        \centering
        \subfloat[Different transient dynamics.]{
        \includegraphics[width=0.9\linewidth]{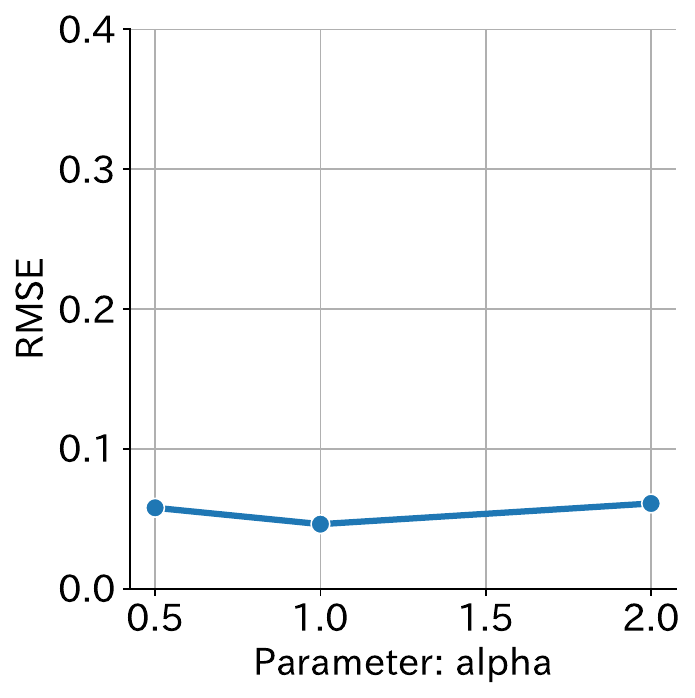}}
    \end{minipage}
    \\
    \begin{minipage}[b]{0.24\textwidth}
        \centering
        \subfloat[Reconstructed (5k)]{\includegraphics[width=0.95\linewidth]{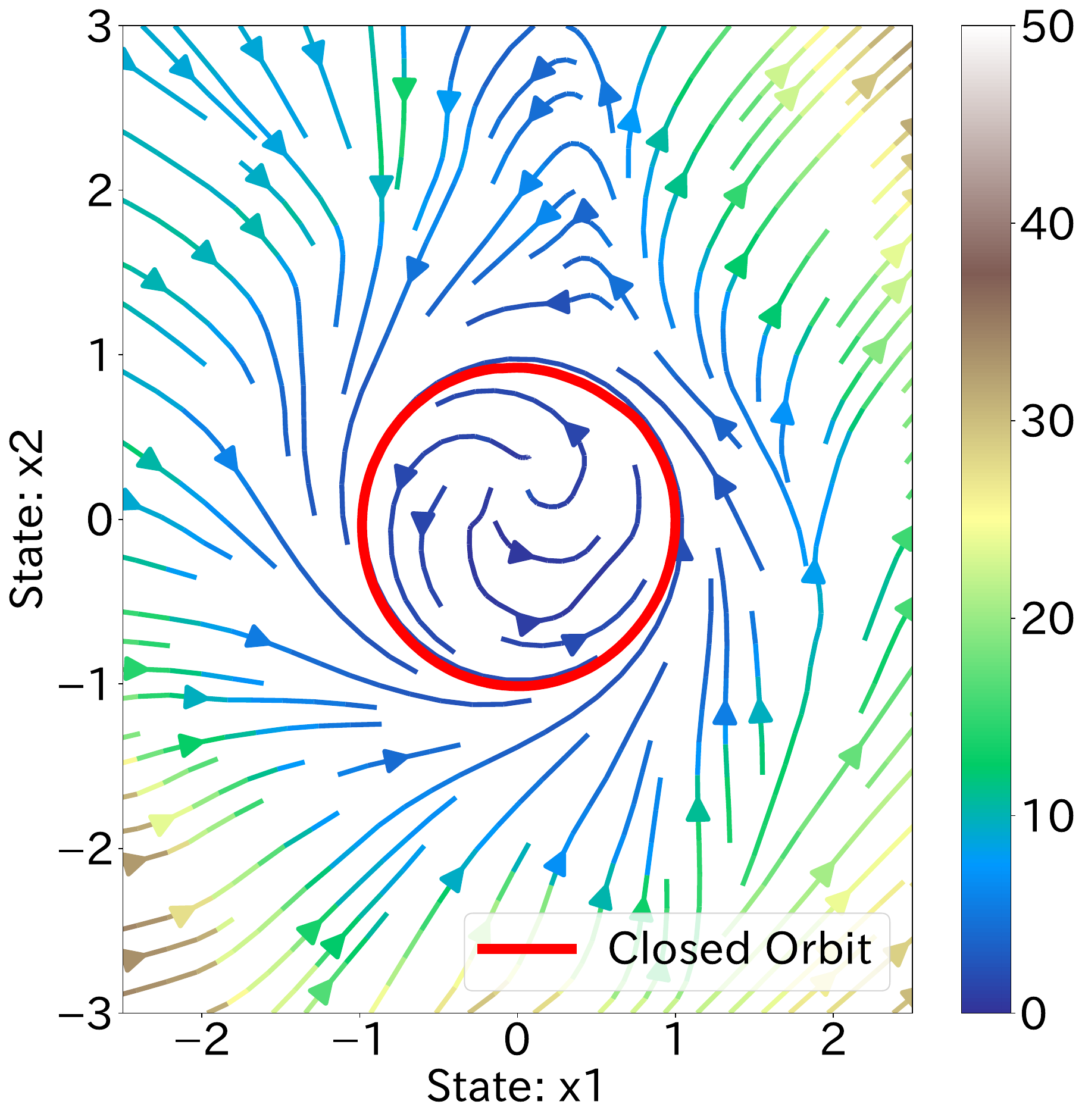}}
    \end{minipage}
    \begin{minipage}[b]{0.24\textwidth}
        \centering
        \subfloat[Reconstructed (10k)]{\includegraphics[width=0.95\linewidth]{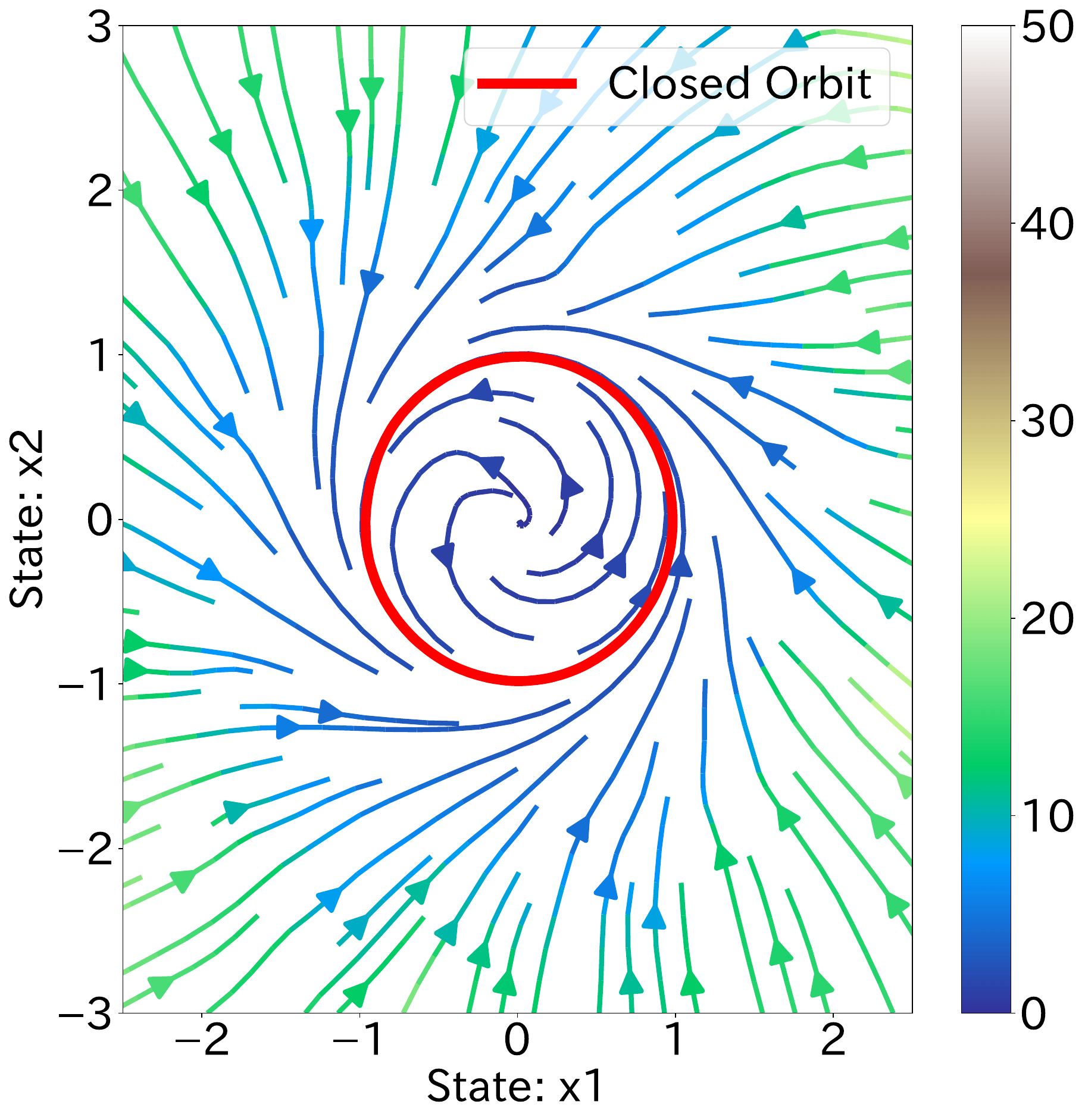}}
    \end{minipage}
    \begin{minipage}[b]{0.24\textwidth}
        \centering
        \subfloat[Reconstructed (50k)]{\includegraphics[width=0.95\linewidth]{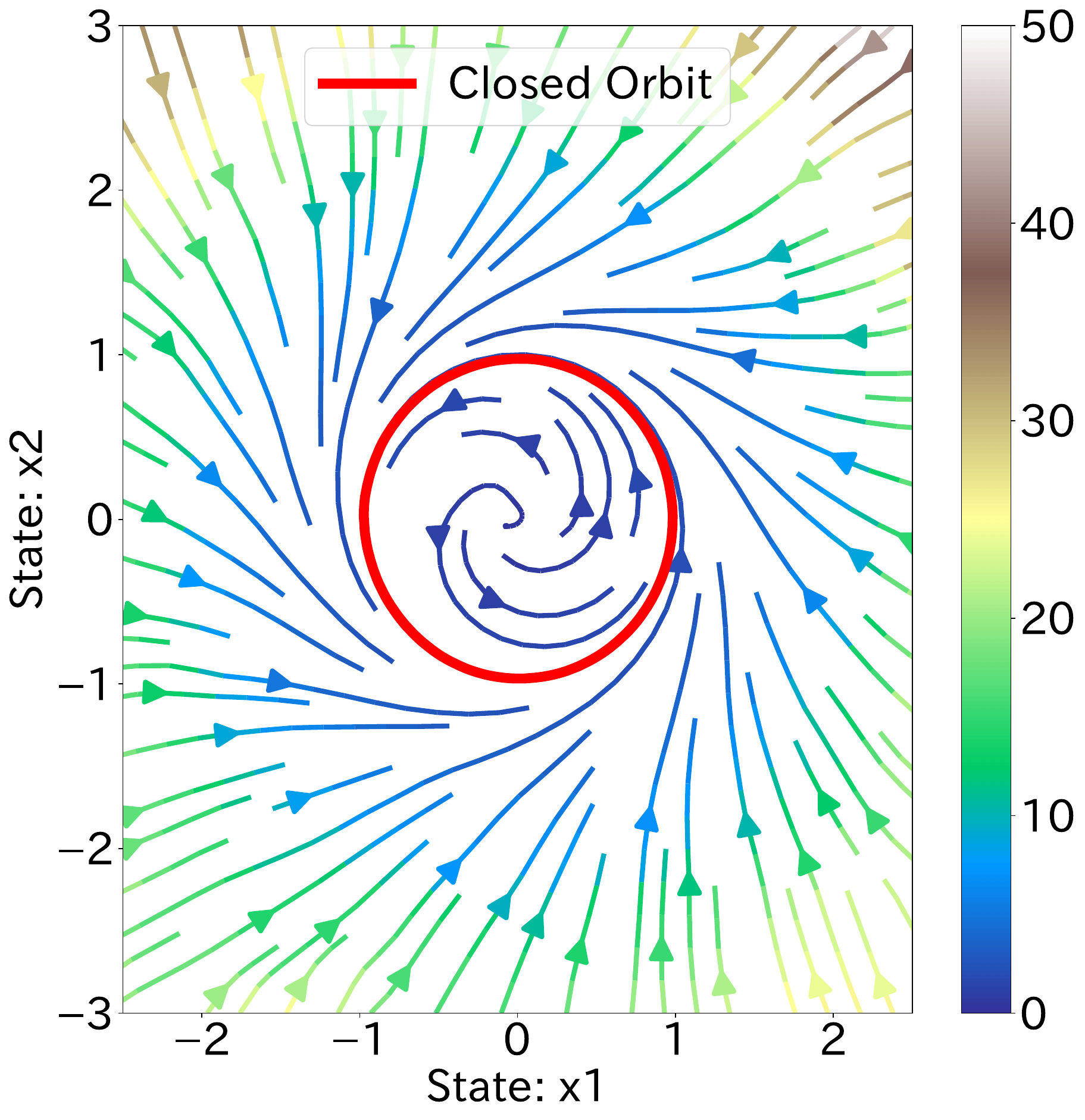}}
    \end{minipage}
    \begin{minipage}[b]{0.24\textwidth}
        \centering
        \subfloat[Reconstructed (100k)]{\includegraphics[width=0.95\linewidth]{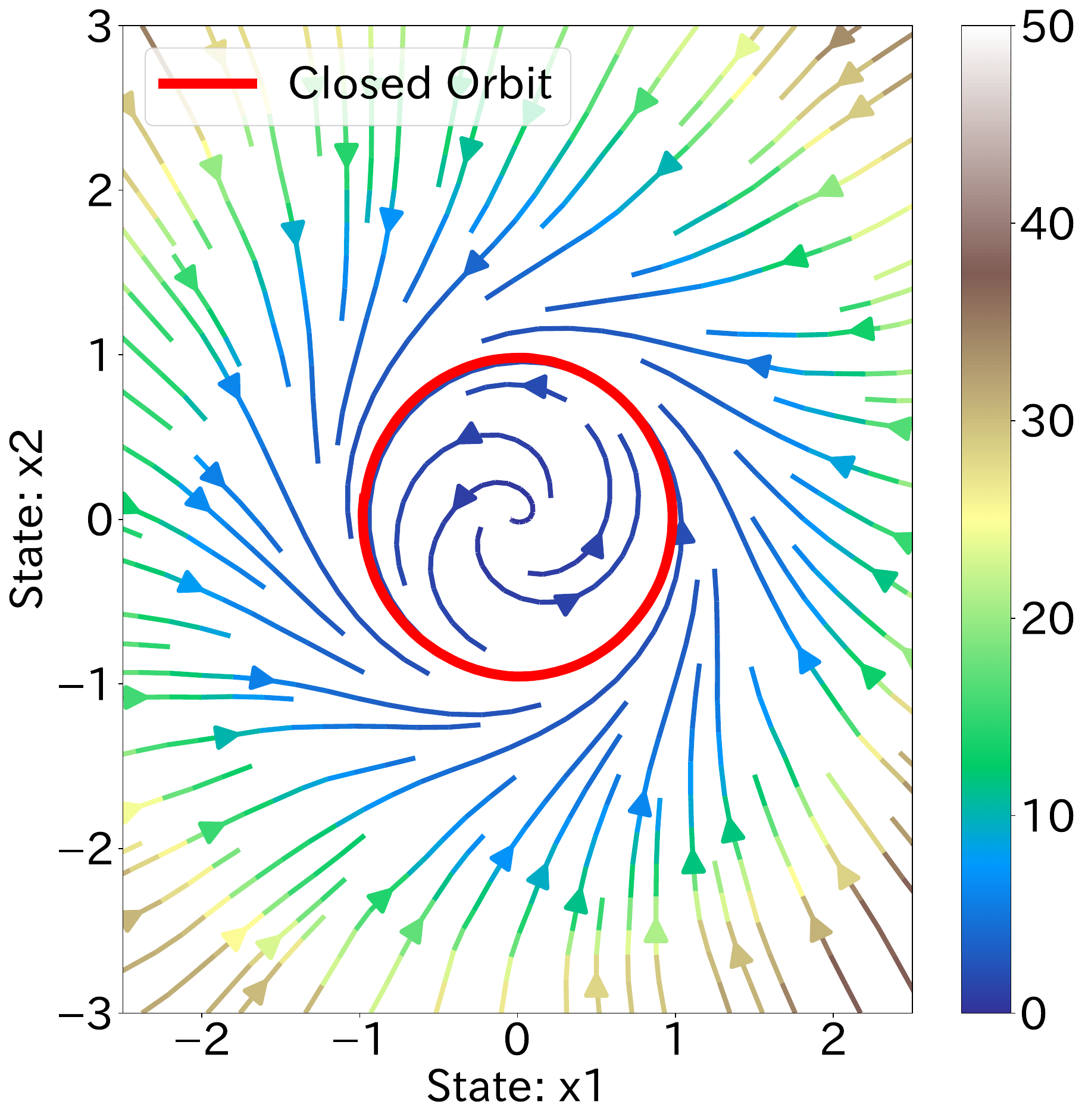}}
    \end{minipage}

    \caption{
    Learning limit cycle attractor.
    (a) Target limit cycle attractor in 2D state space $\boldsymbol{x}=[x_1, x_2]$.
    Color bars show the norm of the vector fields: $\dot{\boldsymbol{x}}=F(\boldsymbol{x})$ and a solid line shows the limit cycle.
    (b) RMSEs with different data sizes. 
    (c) RMSEs with different target dynamics, i.e., different scale parameters $\alpha$. Our proposed methods showed similar approximation performance on the different target dynamics.
    (d) - (g) Reconstructed vector fields corresponding to different data sizes from 5k to 100k time steps. Red line starts after 3.1 s (= 200 timesteps.) to show the approximated limit cycle.
    While transient dynamics reconstruction requires more than 10k steps of data, the limit cycle is well reproduced with 5k steps.
    }
    \label{fig:simple-limit-cycle-figs}
\end{figure}

We generated trajectories from the randomly sampled initial positions using Eq. (\ref{eq:simple-limit-cycle}) and learned the encoder and decoder (Fig. \ref{fig:simple-limit-cycle-figs} (a)). 
Table \ref{tab:parameter-list} lists other learning parameters.
The learned vector field reconstructs the limit cycle and transient dynamics, as shown in Fig. \ref{fig:simple-limit-cycle-figs} (d-g).
We evaluated the root mean square error (RMSE), $\sqrt{1/T\sum_{k=0}^T \|\boldsymbol{x}_k-\zeta(\boldsymbol{z}_k)\|^2_2}$, with 100 trajectories and 400 timesteps per each trajectory as test set; these were not used in the network training process.
Fig. \ref{fig:simple-limit-cycle-figs} (b) shows the RMSE score for each training dataset size.
786 s = (50 k steps) of data collection is sufficient to reconstruct a trajectory under a small enough RMSE score.
\color{\hcolor}
Standard imitation learning approaches use a similar amount of data to train the network model for trajectory reconstruction \cite{pahic2020}.
\color{black}
Fig. \ref{fig:simple-limit-cycle-figs} (c) shows the RMSE score in different scale parameters $\alpha \in \{0.5, 1, 2\}$ where the number of training time steps is 100 k.
The proposed method demonstrated the reconstruction of the limit-cycle dynamical system, which has different transient dynamics.

\begin{table}
    \caption{Parameter List}
    \label{tab:parameter-list}
    \setlength{\tabcolsep}{5pt} 
    \renewcommand{\arraystretch}{1.2}
    \begin{tabular}{cccc}
        Parameters  & Experiment name &  Values&  Notes \\ \hline\hline
        Optimizer   &  &Adam&  \\ \hline
        Learning rate   &  & $10^{-3}$ & \\\hline
        Batch size $B$ & &255 &  \\\hline
        Num. iteration $N$  &   &  5000 &              \\\hline
        Data steps  & Simple limit cycle &  100,000 & 1571 s\\
                    & Lemniscate follow &  400 & 20 s\\
                    & Human-to-Robot & 1208 & 24.2 s  \\\hline
        Split horizon $T$  & Simple limit cycle &  500 & \\
                    & Lemniscate follow &  300 & \\
                    & Human-to-Robot & 295 & $\sim  \frac{T_0}{\Delta T}$, $T_0= \frac{1}{\omega_0}$ s \\hline
        Discount factor $\gamma$    & Simple Limit Cycle &  0.99 & \\
                    & Lemniscate follow &  0.999 & \\
                    & Human-to-Robot & 0.998 & $\sim\exp(-\frac{1}{2T_0})$ \\\hline
        MLP hidden layer size   & Others & [512, 512] & \\
                                & Human-to-Robot & [128, 128] & \\\hline
        Laplace scale $b_f,b_h,b_0$ & &$ 10^{-5} $ & \\\hline
        Intensity parameter $\kappa$ & & 0.5& \\ \hline
        Feedback gain $g$ & Simple limit cycle &  - & \\
                    & Lemniscate follow &  $5 \times 10^{-2}$ / $1 \times 10^{-3}$  & Anomaly / Other.\\
                    & Human-to-Robot & $0.02$  & $\boldsymbol{g}=g/ \omega_0 [\boldsymbol{\omega}, \boldsymbol{\lambda}] $\\\hline
        Discretized step $\Delta T$  & Simple limit cycle&  15.7 ms &  \\
                    &Lemniscate follow & 50 ms &  \\
                    &Human-to-Robot & 20 ms &  \\\hline
        Dim. latent $|z|$  & Simple limit cycle & 2 & \\
                    & Lemniscate follow& 32 & \\
                    & Human-to-Robot & 32 & two phases + 30 amplitude \\\hline
        Characteristic freq. $\omega$ & Simple limit cycle& 2.00 rad/s  &\\
         & Lemniscate follow& 0.2 Hz  &\\
         & Human-to-Robot& [0.17, 0.51] Hz & $=[\omega_0,\omega_1]$ \\\hline
        \multirowcell{3}{
            Characteristic exponents\\
            Range $[\lambda_\mathrm{\min}, \lambda_\mathrm{\max}]$ 
        } & Simple limit cycle & [2.0, 2.0] rad / s  & $=2\alpha$\\
        & Lemniscate follow & [0.039, 6.283] rad / s  & \\
        & Human-to-Robot & [3.19,  9.42] rad / s & \\ \hline
       \hline
    \end{tabular}
\end{table}

\subsection{Lemniscate tracking control}
\begin{figure}
    \centering
    \begin{minipage}[b]{0.35\textwidth}  
        \centering
        \subfloat[Anomaly situation]{\includegraphics[width=\linewidth]{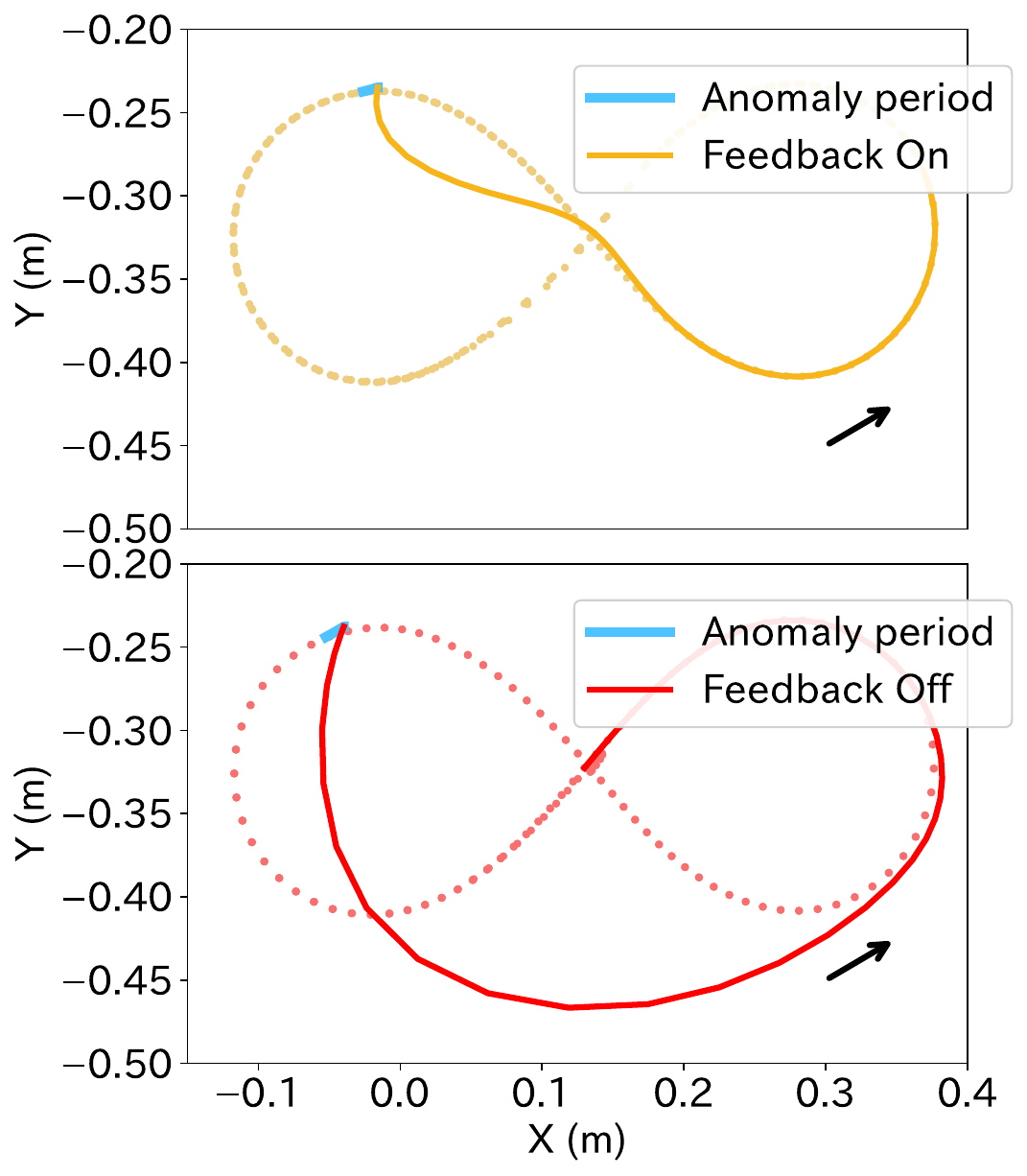}}
    \end{minipage}
    \begin{minipage}[b]{0.4\textwidth}  
        \centering
        \subfloat[Desired and actual X position]{\includegraphics[width=\linewidth]{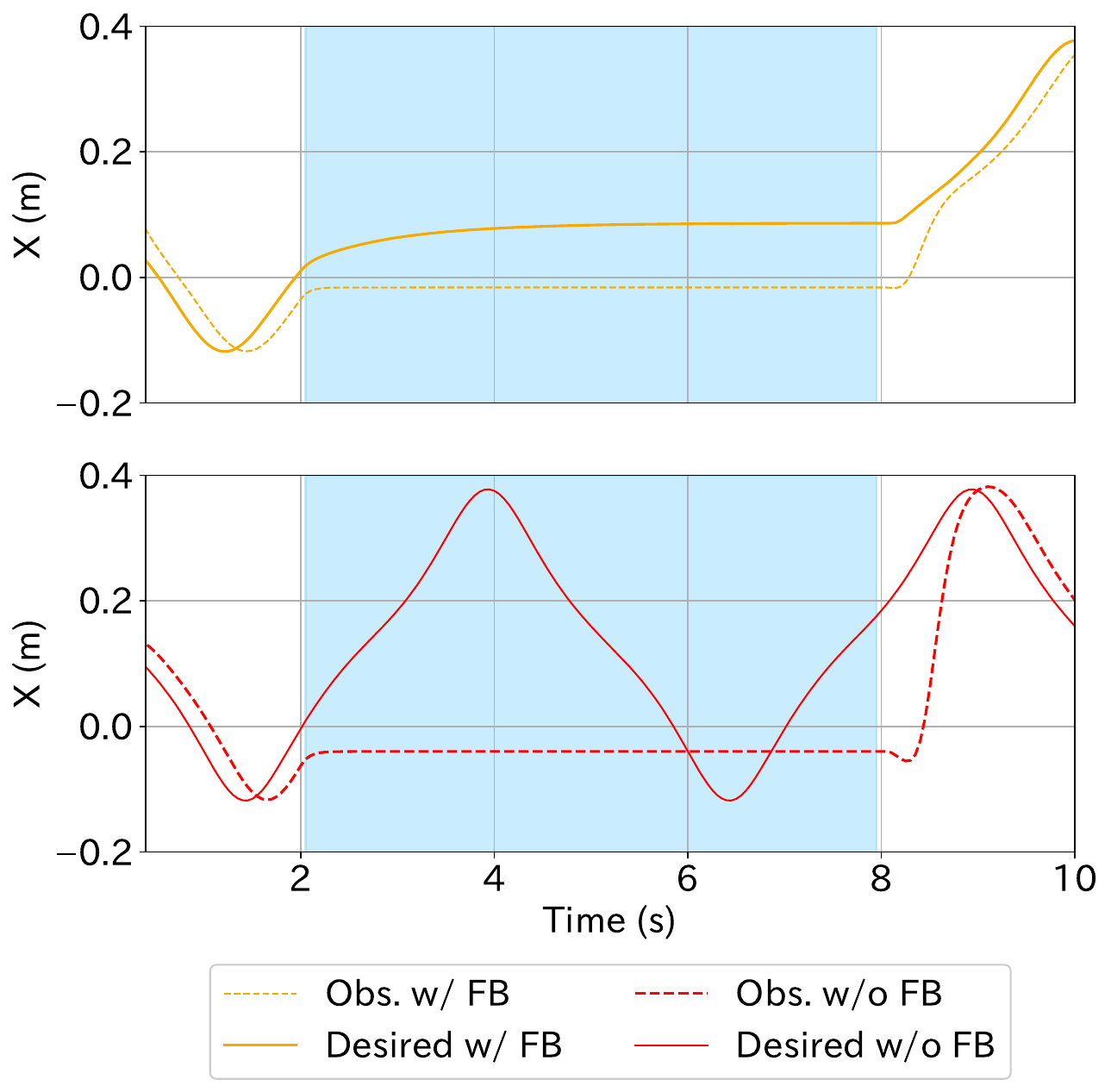}}
    \end{minipage}
    \caption{
    Lemniscate tracking tasks in anomaly situations. Fig. \textbf{(a)}
    Hand position trajectory. Fig. \textbf{(b)} Time series of the X-axis hand position.
    The pale blue area shows the cut control region.
    With feedback, the desired trajectory (yellow line) is corrected to match the hand trajectory, but without feedback, the trajectory (red line) continues, so the hand moves to a distant point when it recovers from an anomaly.
    }
    \label{fig:ur5e-lemniscate-cut-control}
\end{figure}

We considered four transient movement situations to validate our proposed method.
First, we conducted the control task in an anomaly while the control signal suddenly stopped for six seconds.
The desired dynamical system must modulate itself. Otherwise, the robot suddenly accelerates to return to the desired point after recovering from an anomaly.
The feedback mechanism in Eq. (\ref{eq:feedback-system}) was effectively worked to synchronize the latent variable with the real robot state, as shown in Fig. \ref{fig:ur5e-lemniscate-cut-control} (a) and (b). 
Without the feedback term, the robot rapidly attempted to return to the target state after the control was recovered, causing the behavior that deviated significantly from the demonstration. 
Conversely, the proposed method stopped the latent phase based on feedback from the encoded robot state when the control signal was lost. After restoring the control, the robot gently followed the demonstrated trajectory.

\begin{figure}
    \centering
    \begin{minipage}[b]{0.5\textwidth}  
        \centering
        \subfloat[Demonstrated trajectory]{\includegraphics[width=\linewidth]{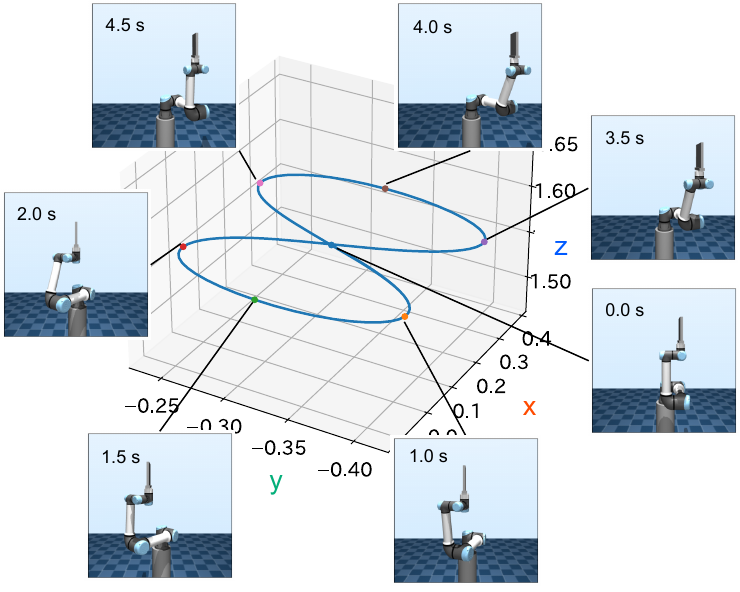}}
    \end{minipage}
    \begin{minipage}[b]{0.45\textwidth}
        \centering
        \begin{minipage}[b]{\linewidth}  
            \centering
            \subfloat[RMSE]{\includegraphics[width=0.9\linewidth]{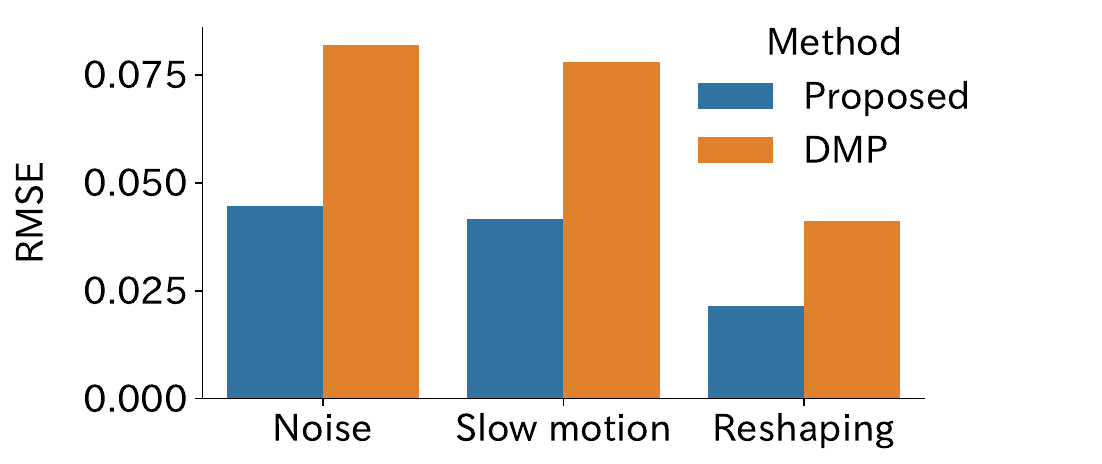}}
        \end{minipage}
    \end{minipage}
    \\
    \centering
    \begin{minipage}[t]{0.3\textwidth}  
        \centering
        \subfloat[Noise: Proposed]{\includegraphics[width=\linewidth]{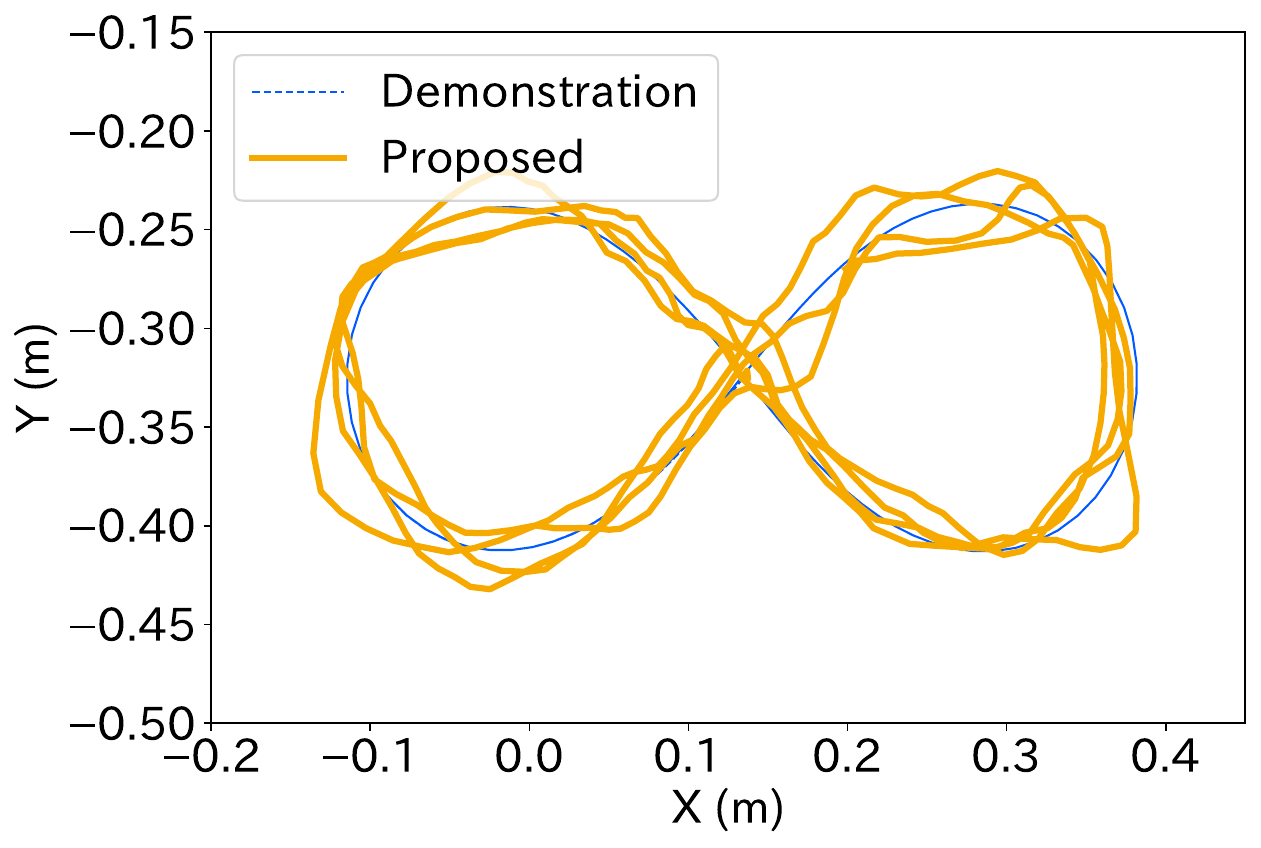}}
    \end{minipage}
    \begin{minipage}[t]{0.3\textwidth} 
        \centering
        \subfloat[Slow motion: Proposed]{\includegraphics[width=\linewidth]{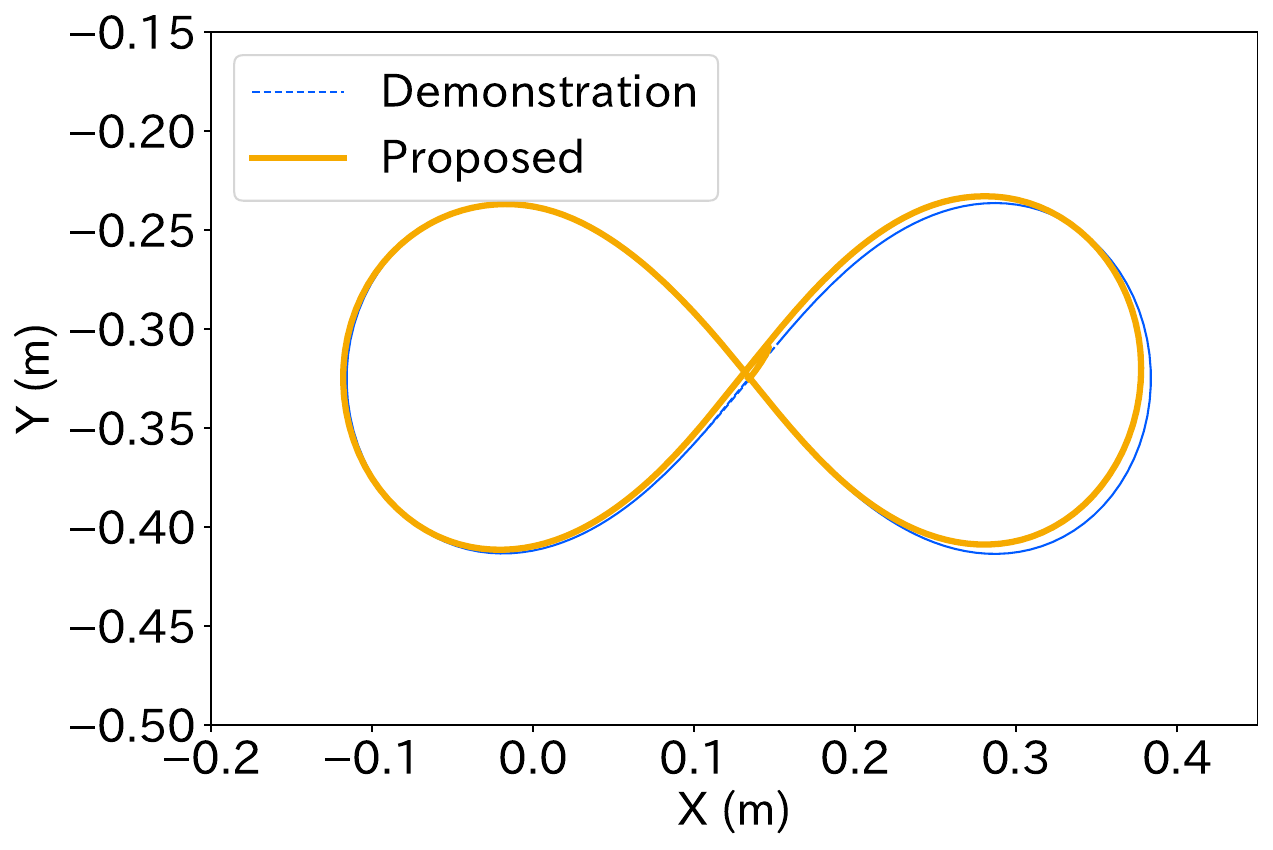}}
        \label{fig:slow-motion-proposed}  
    \end{minipage}
    \begin{minipage}[t]{0.3\textwidth} 
        \centering
        \subfloat[Reshaping: Proposed]{\includegraphics[width=\linewidth]{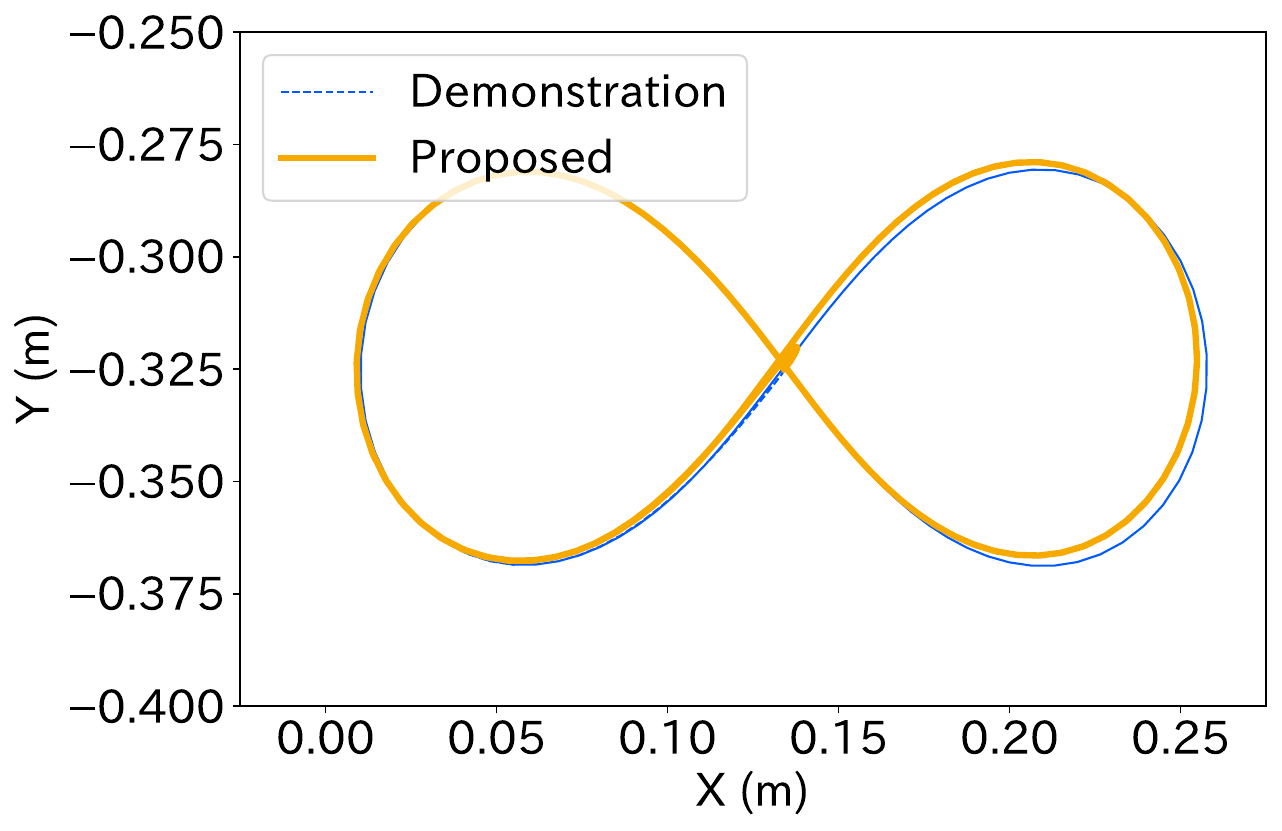}}
        \label{fig:reshaping-proposed}
    \end{minipage}
    \\
    \begin{minipage}[b]{0.3\textwidth}  
        \centering
        \subfloat[Noise: DMP]{\includegraphics[width=\linewidth]{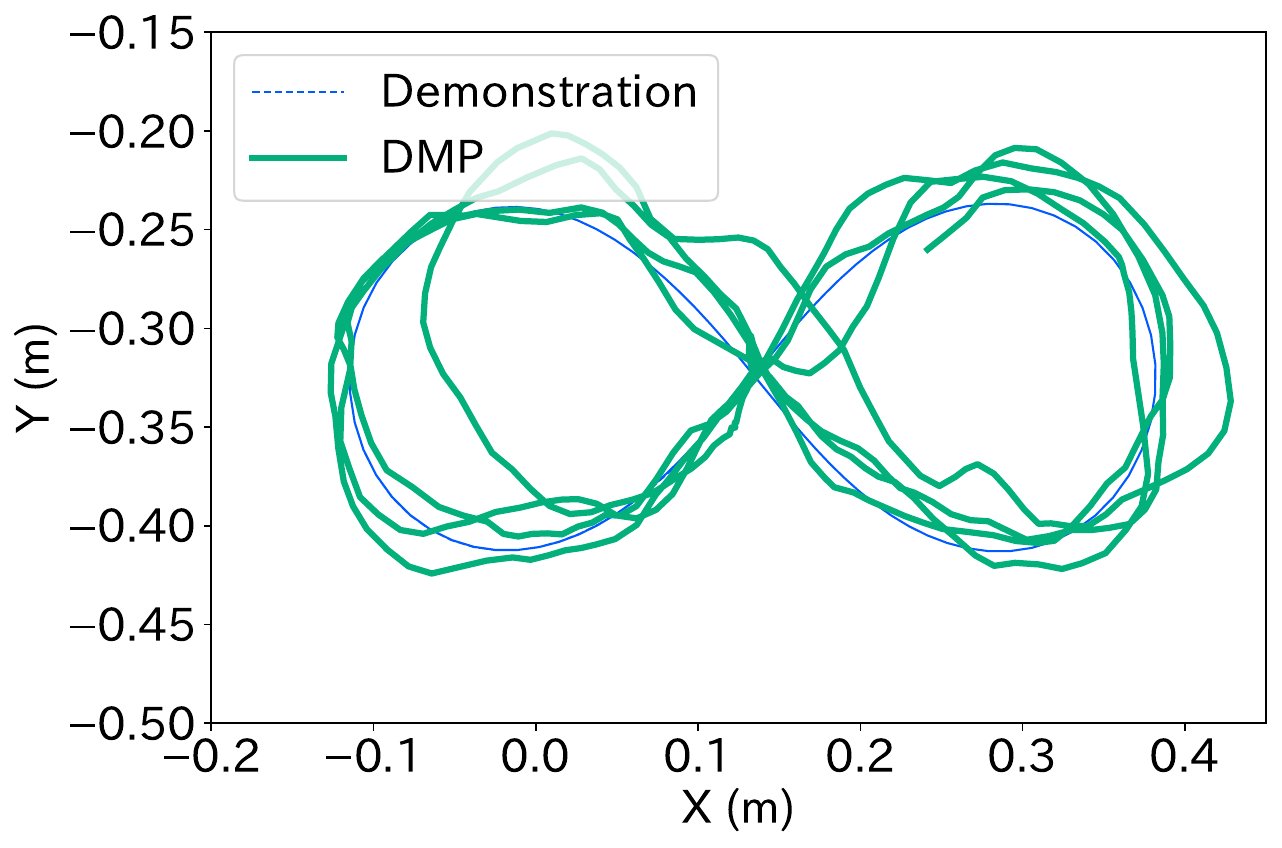}}
    \end{minipage}
    \begin{minipage}[b]{0.3\textwidth}  
        \centering
        \subfloat[Slow motion: DMP]{\includegraphics[width=\linewidth]{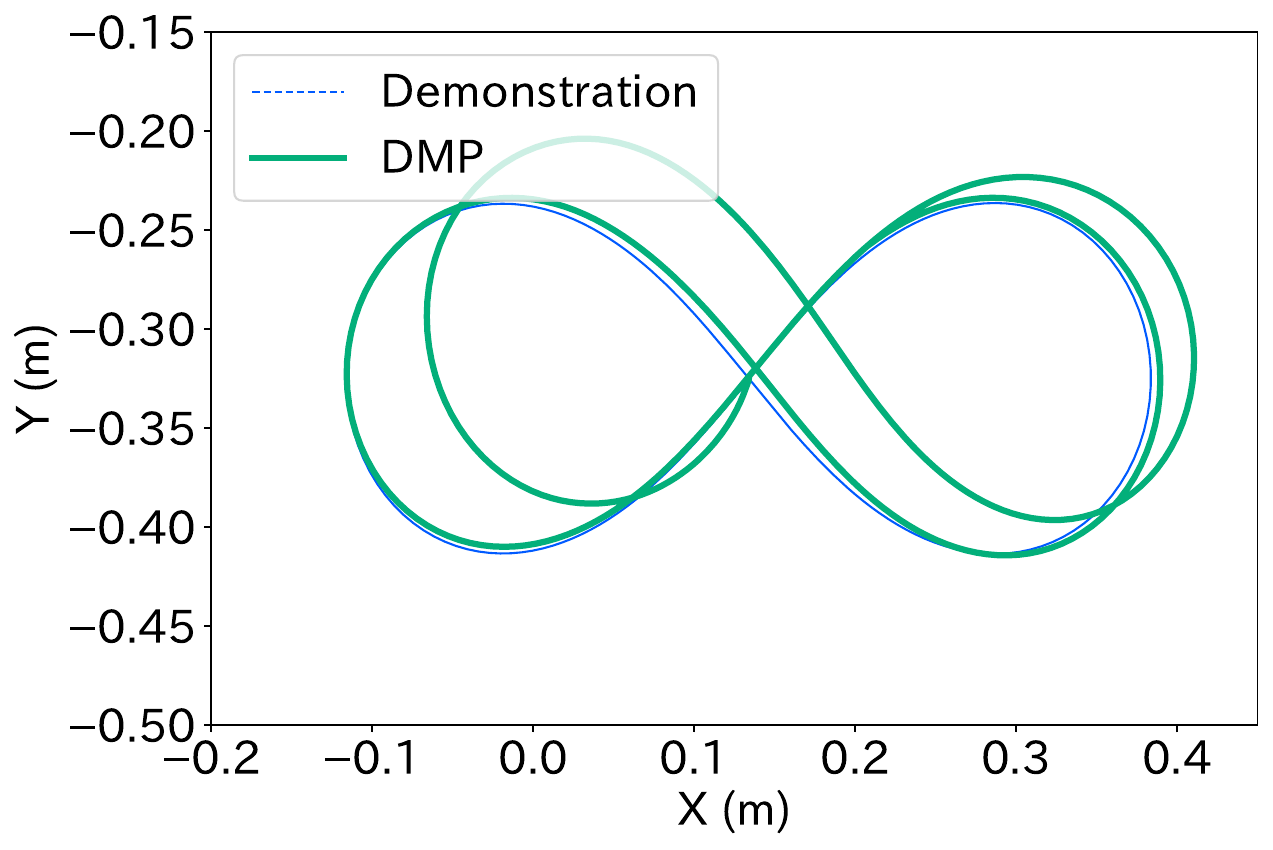}}
        \label{fig:slow-motion-dmp}
    \end{minipage}
    \begin{minipage}[b]{0.3\textwidth}  
        \centering
        \subfloat[Reshaping: DMP]{\includegraphics[width=\linewidth]{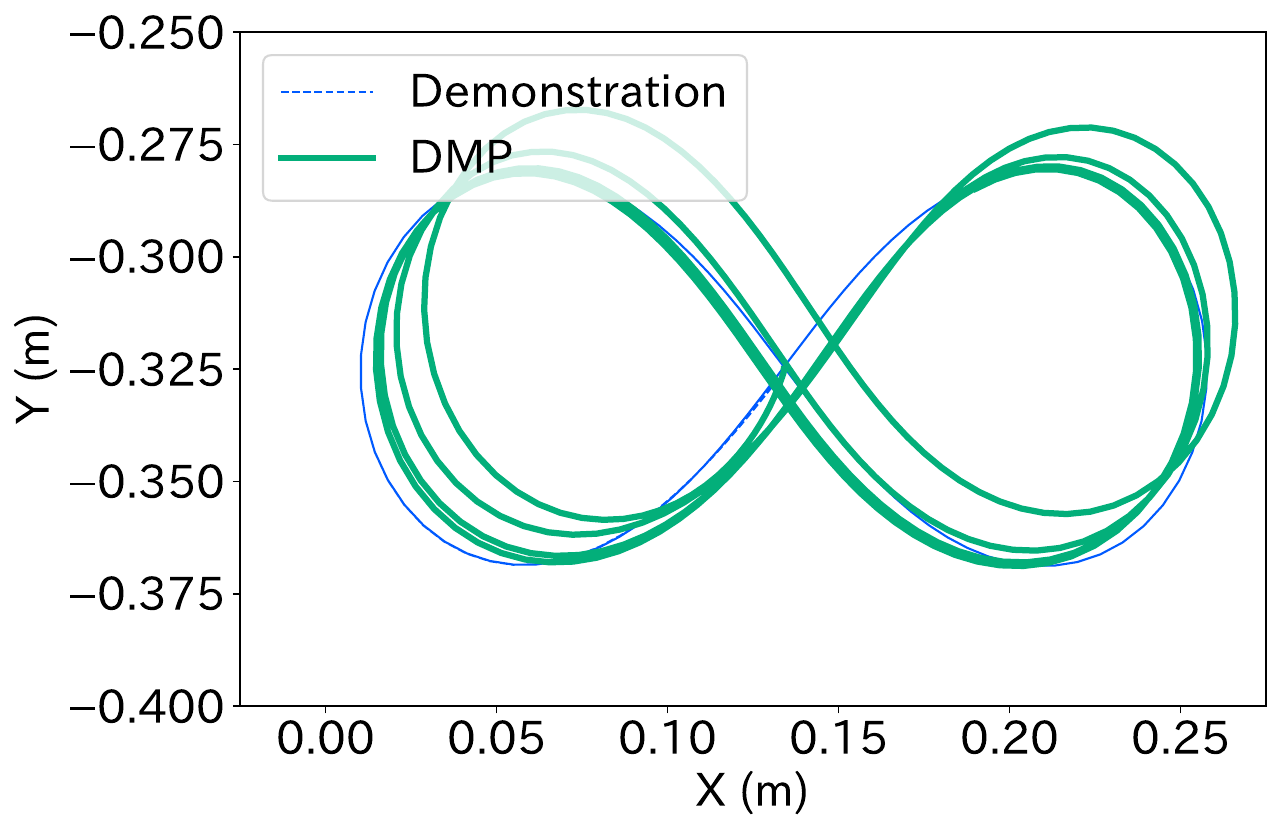}}
        \label{fig:reshaping-dmp}
    \end{minipage}
    \caption{
    Lemniscate tracking tasks: force noise injection Fig. \textbf{(b, c, f)}, slow motion \textbf{(b, d, g)}, and trajectory reshaping Fig. \textbf{(b, e, h)}.
    The demonstration data was generated in the lemniscate curve with a constant frequency of 0.2 \nolinebreak Hz for 20 s.
    }
    \label{fig:ur5e-lemniscate}
\end{figure}

Second, we compared the robustness against the force noise from isotropic Gaussian distribution with mean 0 N and standard deviation 100 N at the hand position.
Because the proposed method models the amplitude components dynamics, it can be generated to converge quickly behavior to the demonstrated trajectory, compared to the DMP generated dynamics, as shown in Fig. \ref{fig:ur5e-lemniscate} (c) and (f). The RMSE score between the proposed method and the demonstration was superior to that of the DMPs, as shown in Fig. \nolinebreak\ref{fig:ur5e-lemniscate} (b).

Third, we considered the slow-motion task that forces the robot to decrease the tracking speed in half.
Because the time transition part and observation space are entirely separated, the proposed method makes it straightforward to scale the phase of the limit cycle.
The trajectory-following shape dynamics are shown in Figs. \ref{fig:ur5e-lemniscate} (d) and \ref{fig:ur5e-lemniscate} (g). The waveform collapses along the slow phase because the phase and amplitude components are not separated in DMP.

Fourth, we considered the reshaping task that imposes the robot shrinkage of the size of the limit cycle in half.
Here, the proposed method shrinks the waveform's shape while preserving the phase, whereas the DMP is slightly delayed in phase, as shown in Fig. \ref{fig:ur5e-lemniscate} (e) and Fig. \ref{fig:ur5e-lemniscate} (h).
The RMSE scores of the third and fourth tasks between the proposed method and the demonstration were superior to that of the DMPs, as shown in Fig. \ref{fig:ur5e-lemniscate} (b).
Although slow-motion and reshaping tasks were studied based on orientation-preserving homeomorphism proposed by \cite{ijspeert2013}, it is difficult to reproduce the same in the dynamic domain because the multi-DoF dynamics has a complex interaction. 
Nonetheless, our approach based on the phase-amplitude equation can be applied to the scaling task with orientation-preserving homeomorphism.

\begin{table}
    \centering
    \caption{Observable $x$ List}
    \renewcommand{\arraystretch}{1.2}
    \setlength{\tabcolsep}{10pt} 
    \label{tab:observables}
    \begin{tabular}{c}
        \hline \hline
        Simple Limit Cycle  \\ \hline
        $[x_1,x_2]\in \mathbb{R}^2$ 
        \\ \hline\hline
        Lemniscate Following \\ \hline
        Tool Center Position in $\mathbb{R}^{3}$\\ \hline
        Tool Center Velocity in $\mathbb{R}^{3}$\\ 
        \hline \hline
        Human-to-Robot \\ \hline
        The head and tail position of the rod  in $\mathbb{R}^6$ \\ \hline
        The head and tail velocity of the rod in $\mathbb{R}^6$\\\hline \hline
    \end{tabular}
\end{table}

\subsection{Human movement imitation}
\label{sec:human-movement-imitation-result}
\begin{figure}
    \centering
    \begin{minipage}[b]{0.9\linewidth}  
        \subfloat[Imiteted baton waving movement.]{\includegraphics[width=\linewidth]{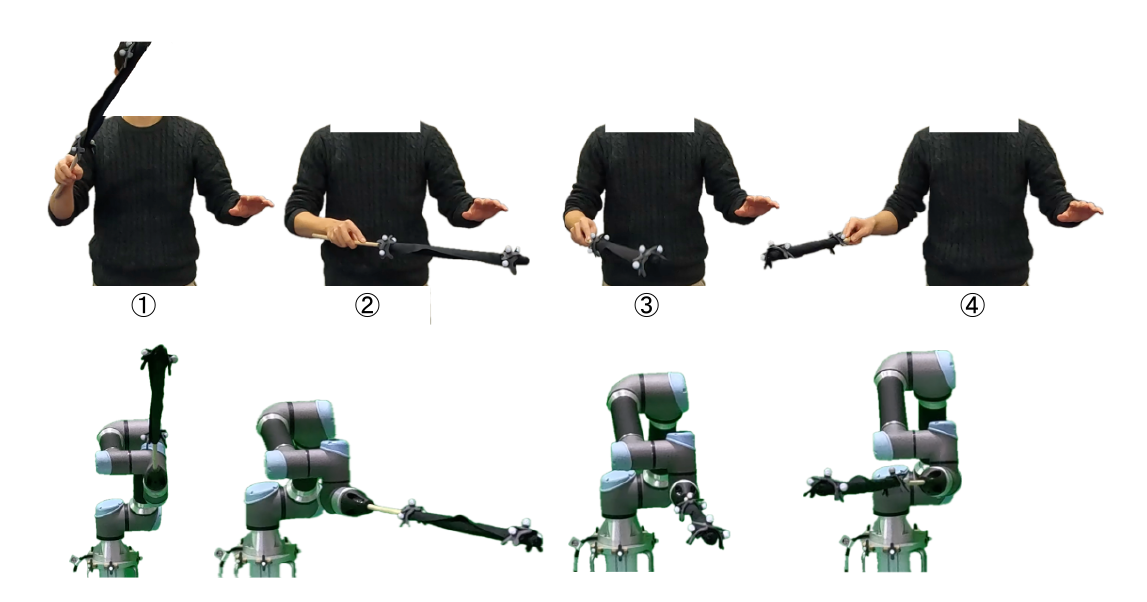}}
    \end{minipage}
    \\
     \begin{minipage}[c]{0.4\linewidth}  
        \centering
        \subfloat[Experimental setup.]{\includegraphics[width=\linewidth]{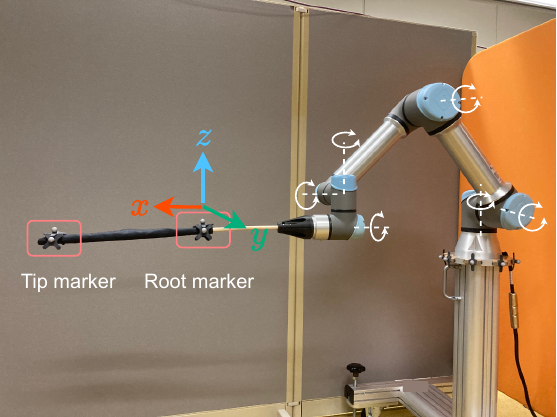}}
    \end{minipage}
    \begin{minipage}[c]{0.4\linewidth}  
        \centering
        \subfloat[Generated baton waving trajectories.]{\includegraphics[width=\linewidth]{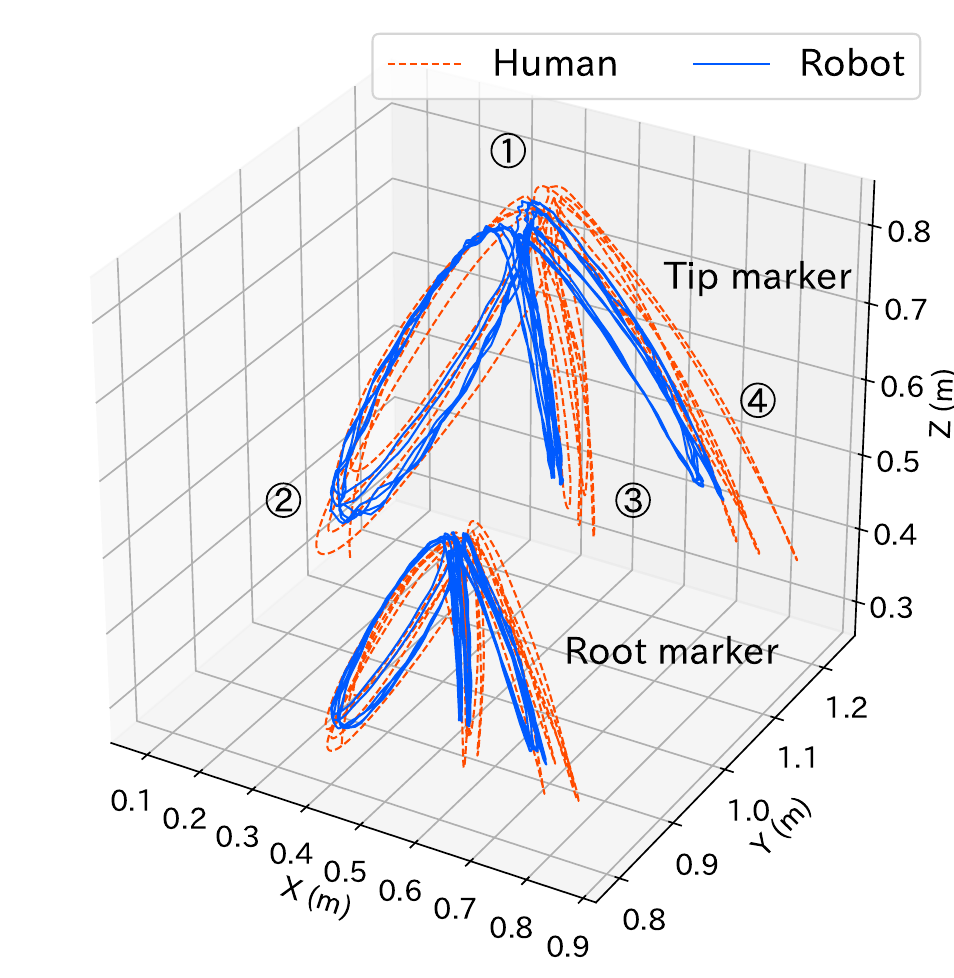}}
    \end{minipage}
    \\
    \begin{minipage}[c]{0.55\linewidth}  
        \centering
        \subfloat[Predicted trajectory ]{\includegraphics[width=\linewidth]{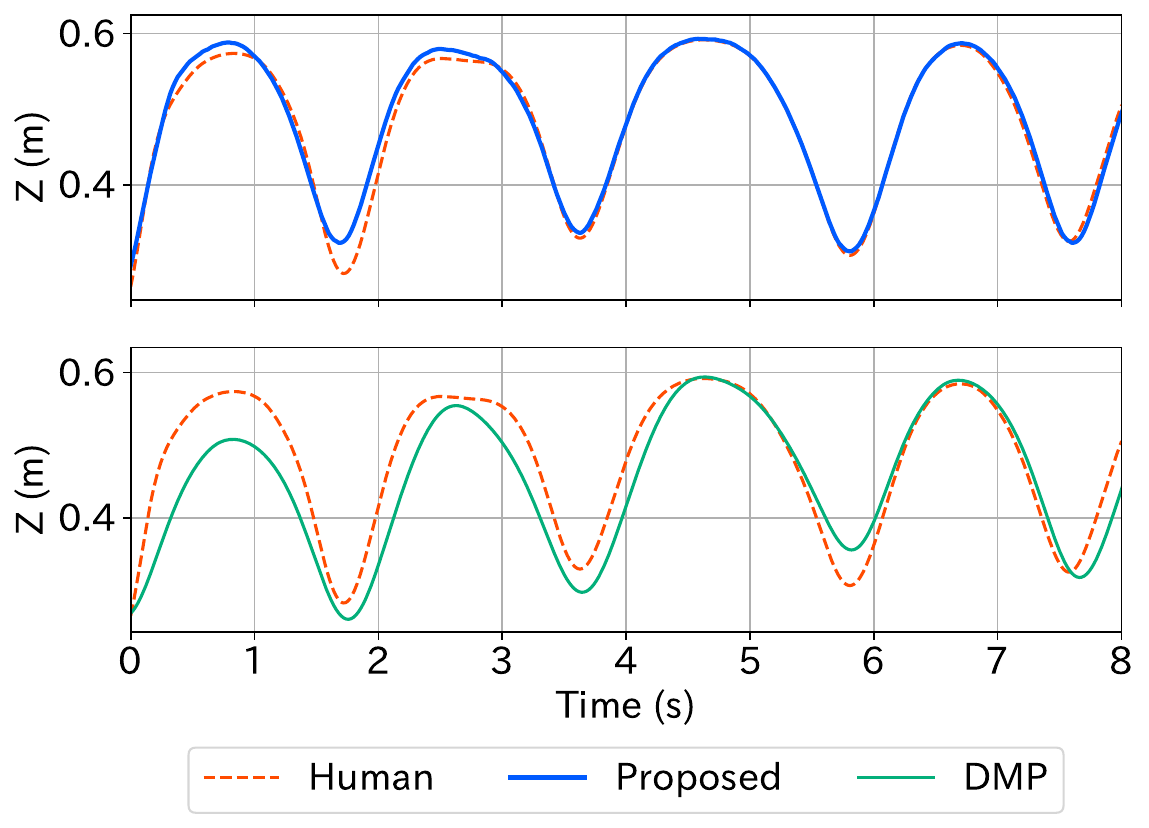}}
    \end{minipage}
    \caption{
    Imitation learning experiment.
    a) Baton is waving with three beats and 30 tempos per 60 s.
    b) Robot arm system with markers to capture movements.
    c) Generated baton motions by a human and a robot arm. Marker positions attached to the tip and root of the baton are plotted.
    The arm robot successfully imitated the demonstrated human baton-waving motion.
    d) Trajectory prediction performances of our proposed method and DMP. Our proposed method showed more accurate prediction performance than DMP.
    e) Real robot trajectory with/without interactive feedback in control cut anomaly. Without feedback, the hand position becomes discontinuous when the robot recovers from an anomaly. With feedback, the robot maintains the desired motion.
    }
    \label{fig:human-to-robot-result}
\end{figure}

Figure \ref{fig:human-to-robot-result} (a) depicted the snapshot of the imitated movement when the robot follows the decoded trajectory generated by the latent dynamics learned from the human baton waving motion. 
The learned model correctly reconstructed the human trajectory despite the original human behavior being perturbed around a limited cycle. The reconstructed trajectory was tracked by the arm robot using a servo controller. The resulting movement of the robot was measured by a motion capture system using markers put on the robot (Fig. \ref{fig:human-to-robot-result} (b))
As plotted in Fig. \ref{fig:human-to-robot-result} (c), the robot arm successfully imitated the human baton-waving motion.

We compared the predicted performance of the proposed method with DMP to validate the model performance in predicting human data.
The proposed method reconstructs the two-dimensional torus behavior of the baton-waving motion. Contrary, DMP is limited in its representation of the dynamical system as shown and, therefore, does not reproduce the trajectory of the limit cycle in Fig. \ref{fig:human-to-robot-result} (d). The RMSE score of Proposed was 0.080, and the DMP was 0.094.

\begin{figure}
    \centering
    \begin{minipage}[c]{0.45\linewidth}  
        \centering
        \subfloat[With interactive feedback.]{\includegraphics[width=\linewidth]{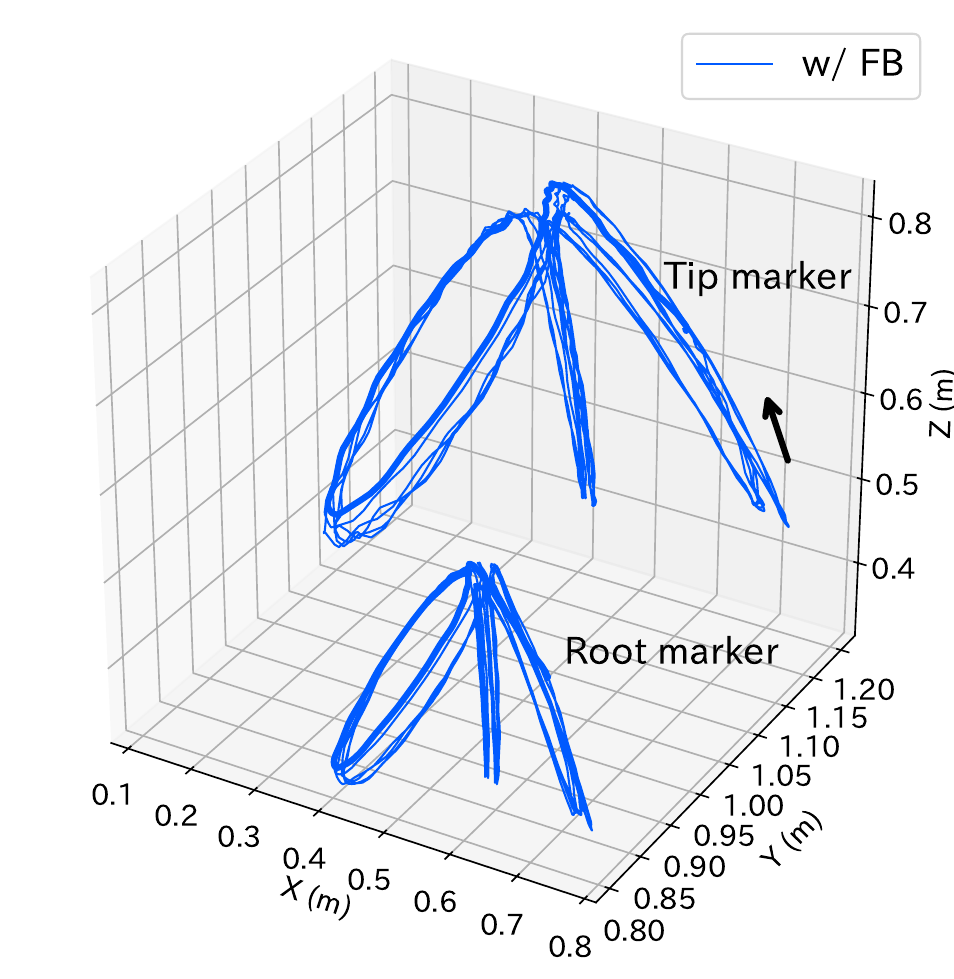}}
    \end{minipage}
    \begin{minipage}[c]{0.45\linewidth}  
        \centering
        \subfloat[Without interactive feedback.]{\includegraphics[width=\linewidth]{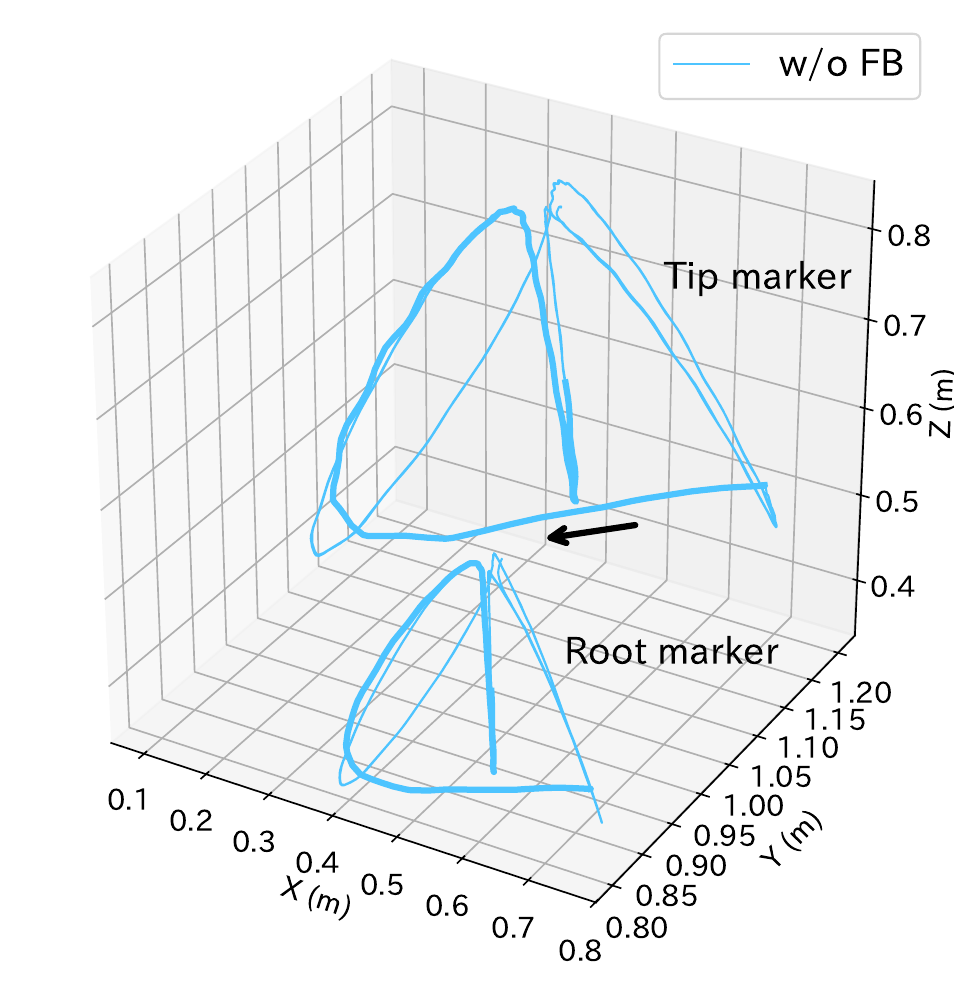}}
    \end{minipage}
    \caption{
    Real robot trajectory with/without interactive feedback in control cut anomaly. 
    a) With feedback, the robot maintains the desired motion.
    b) Without feedback, the desired hand position becomes discontinuous when the robot recovers from an anomaly. 
    }
    \label{fig:human-to-robot-result-with-cutcntrl}
\end{figure}

To demonstrate the effectiveness of the feedback term on a real robot, we induced the control anomaly on the 6DOF robot, as shown in \textcolor{\hcolor}{Fig. \ref{fig:human-to-robot-result-with-cutcntrl}}. The feedback controller on the robot was terminated and subsequently reactivated. In the absence of feedback, the desired trajectory was modified to a substantially different state, resulting in a significant change in the robot movement. In contrast, when feedback was available, the robot was able to maintain the desired trajectory and continue tracking it.




\section{Conclusion}
We introduced a novel framework for learning forward and inverse mappings to phase-amplitude dynamics, enabling imitation learning in movement tasks from human to robot. Additionally, we developed an interactive feedback control mechanism to cope with external disturbance or anomaly situations. Our proposed method was validated through two simulated experiments and a real imitation learning experiment. Consequently, the acquired latent dynamic model by the proposed method successfully reproduced the limit cycle and the transient dynamics; moreover, a 6DoF arm robot replicated the human rhythmic motion. 


With the proposed method, our learning system cannot imitate the human behavior of switching between multiple attractors. For example, in a complex assembly task, a human would switch attractors sequentially while having the left and right arms generate separate attractors. 
To cope with this kind of situation, as a future study, we consider training multiple networks to integrate different attractor dynamics.

\section*{Appendix}
\subsection*{Loss Function Derivation}
We derive the absolute error loss from the KL divergence in Eqs. (\ref{eq:val-prob-1})-(\ref{eq:targ-prob-4}).
The loss functions $\mathrm{KL}[q_1|p_1]$, $\mathrm{KL}[q_1|p_2]$, $\mathrm{KL}[q_2|p_3]$, $\mathrm{KL}[q_2|p_4]$ were derived using an approach similar to that in Section \ref{sec:Encoder-Decoder-Learning} as follows:
The first KL divergence $\mathrm{KL}[q_1|p_1]$ is equivalent to Eq. \eqref{eq:kl-ae-expantion}, except for the $\beta$ parameter of VAE.
Therefore, the derivation is also the same.
\begin{align} 
 \mathrm{KL}[q_1|p_1] &= \mathbb{E}_{q_1} \left [-\beta \log p(\boldsymbol{z}_0) +\sum_{k=0}^{T}-\log d(\boldsymbol{x}_k|\boldsymbol{z}_k) \right] + \mathrm{const} \nonumber \\
 &= \mathbb{E}_{q_1} \left [\frac{\beta}{b_0} |\boldsymbol{z}_0| +\sum_{k=0}^{T} \left|\boldsymbol{x}_k - \zeta(\boldsymbol{z}_k)\right| \right ]+ \mathrm{const}.
\end{align}
The second KL loss $\mathrm{KL}[q_1|p_2]$ is derived as follows:
\begin{align} 
\mathrm{KL}[q_1|p_2] &= \mathbb{E}_{q_1} \left [\sum_{k=1}^{T}\log m_1(\boldsymbol{z}_k|\boldsymbol{z}_0)
    -\sum_{k=1}^{T}\log m_2(\boldsymbol{z}_k|\boldsymbol{z}_0, \boldsymbol{x}_k)
\right] + \mathrm{const} \nonumber\\
&=  \mathbb{E}_{q_1} \left [
    \sum_{k=1}^{T} \left | \boldsymbol{z}_k - f(\boldsymbol{z}_0, \boldsymbol{x}_k, k) \right |
\right] + \mathrm{const} \nonumber \\
&\leq \mathbb{E}_{q_1} \left [
    \sum_{k=1}^{T} \kappa \left | h(\boldsymbol{x}_k) - f(\boldsymbol{z}_0, k) \right | + \left|\boldsymbol{\epsilon}_k\right|
\right] + \mathrm{const}.
\end{align}
Note that the Laplace distribution entropy $\mathbb{E}_{m_1}[-\log m_1] = 1 + \log(2b_f)$ is independent of the location parameter $f(z_0, k)$.
Moreover, the last inequality is based on the following formulation:
\begin{align} 
    \boldsymbol{z_k} - f(\boldsymbol{z}_0, \boldsymbol{x}_k,k) &= f(\boldsymbol{z}_0, k) + \boldsymbol{\epsilon}_k - f(\boldsymbol{z}_0, k) - \kappa (h(\boldsymbol{x}_k) - f(\boldsymbol{z}_0, k)) \nonumber\\
        &= -\kappa \left (h(\boldsymbol{x}_k) - f(\boldsymbol{z}_0, k) \right)+ \boldsymbol{\epsilon}_k  \nonumber \\
    |\boldsymbol{z}_k-f(\boldsymbol{z}_0, \boldsymbol{x}_k,k)|& \leq \kappa |h(\boldsymbol{x}_k) - f(\boldsymbol{z}_0, k)| + |\boldsymbol{\epsilon}_k|.
\end{align} 
The third KL loss can be expanded as follows:
\begin{align}
    \mathrm{KL}[q_2|p_3]&= \mathbb{E}_{q_2} \left [
    - \beta \log p(\boldsymbol{z}_0) 
    - \sum_{k=1}^{T} \log m'_1(\boldsymbol{z}_{k}|\boldsymbol{z}_{k-1}) 
    -\sum_{k=0}^{T}\log d(\boldsymbol{x}_k|\boldsymbol{z}_k)
    \right]  \nonumber\\
    &\qquad + \mathrm{const}  \nonumber \\
    &= \mathbb{E}_{q_2} \left [
     \beta \frac{|\boldsymbol{z}_0|}{b_0}
    + \sum_{k=1}^{T} |\boldsymbol{z}_{k} - f(\boldsymbol{z}_{k-1}, 1)|
    +\sum_{k=0}^{T} |\boldsymbol{x}_k - \zeta(\boldsymbol{z}_k)|
    \right] \nonumber \\
    & \qquad + \mathrm{const}  \nonumber \\
    &\leq \mathbb{E}_{q_2} \left [
    \sum_{k=1}^{T} |h(\boldsymbol{x}_{k}) - f(\boldsymbol{z}_{k-1}, 1)| 
    +\sum_{k=0}^{T} |\boldsymbol{x}_k - \zeta(\boldsymbol{z}_k)|
    \right]\nonumber\\
    & \qquad + \mathbb{E}_{q_2} \left [
     \beta \frac{|\boldsymbol{z}_0|}{b_0}
    + \sum_{k=1}^{T} |\boldsymbol{\epsilon}_k| 
    \right] + \mathrm{const}.
\end{align}
Here, the entropy term $\mathbb{E}_{e}[-\log e] = 1 + \log(2b_h)$ is also independent of the encoder $h$.

The last inequality is based on the following formulation:
\begin{align} 
    \boldsymbol{z}_k - f(\boldsymbol{z}_{k-1}, 1) &= h(\boldsymbol{x}_k) + \boldsymbol{\epsilon}_k - f(\boldsymbol{z}_{k-1}, 1)\nonumber\\
    |\boldsymbol{z}_k - f(\boldsymbol{z}_{k-1}, 1)|& \leq |h(\boldsymbol{x}_k) - f(\boldsymbol{z}_{k-1}, 1)| + |\boldsymbol{\epsilon}_k|.
\end{align}
The fourth KL Loss function is derived as follows.
\begin{align}
    \mathrm{KL}[q_2|p_4] &=\mathbb{E}_{q_2} \left [- \sum_{k=1}^{T} \log m_2(\boldsymbol{z}_k|\boldsymbol{z}_{k-1},x_k) \right]  + \mathrm{const} \nonumber \\
    &= \mathbb{E}_{q_2} \left [\sum_{k=1}^{T} |\boldsymbol{z}_k - f(\boldsymbol{z}_{k-1}, \boldsymbol{x}_k, 1)| \right] + \mathrm{const} \nonumber \\
    &\leq\mathbb{E}_{q_2} \left [\sum_{k=1}^{T} (1-\kappa)|h(\boldsymbol{x}_k) - f(\boldsymbol{z}_{k-1}, 1)| + |\boldsymbol{\epsilon}_k| \right] + \mathrm{const}.
\end{align}
We repeatedly use the encoder’s entropy independence of the location parameter.
The final inequality is based on the following formulation:
\begin{align}
    \boldsymbol{z}_k - f(\boldsymbol{z}_{k-1}, \boldsymbol{x}_k, 1) &= \boldsymbol{z}_k - f(\boldsymbol{z}_{k-1}, 1) - \kappa(h(\boldsymbol{x}_k) - f(\boldsymbol{z}_{k-1}, 1))\nonumber \\
    &= h(\boldsymbol{x}_k)+ \boldsymbol{\epsilon}_k  - f(\boldsymbol{z}_{k-1}, 1) - \kappa(h(\boldsymbol{x}_k) - f(\boldsymbol{z}_{k-1}, 1)) \nonumber \\
    &=(1 -\kappa)(h(\boldsymbol{x}_k) - f(\boldsymbol{z}_{k-1}, 1)) + \boldsymbol{\epsilon}_k \nonumber \\
    &\leq (1 - \kappa) |h(\boldsymbol{x}_k) - f(\boldsymbol{z}_{k-1}, 1)| + |\boldsymbol{\epsilon}_k|.
\end{align}

\section{Acknowledgments}
This study used several open-source software projects, such as PyTorch \cite{paszke2019}, MuJoCo Menagerie \cite{Zakka2022Menagerie}, dm-control \cite{tunyasuvunakool2020}, and DMP code collection \cite{DMPCodesCollection}. We are grateful to the project contributors for providing valuable machine-learning tools.

\section{Funding}
This work was supported by JSPS; under KAKENHI [Grant number JP19J22987, JP22H04998, JP23K24925], NEDO under [project JPNP20006]; JST under Moonshot R\&D program [Grant number JPMJMS223B-3], and Tateisi Science and Technology Foundation.

\section{Disclosure statement}
No potential conflict of interest was reported by the author(s).

\bibliographystyle{tfnlm}
\bibliography{DynamicsRetargeting}

\end{document}